\documentclass[10pt,a4paper]{article}
\usepackage[latin1]{inputenc} %Accentos
\usepackage[spanish,english,es-nolists]{babel} %Castellano
\usepackage[T1]{fontenc}
\usepackage{graphicx}
\usepackage{amssymb}
\usepackage{siunitx}
\usepackage{subfigure}
\usepackage{hyperref}       % hyperlinks
\usepackage{url}            % simple URL typesetting
\usepackage{booktabs}       % professional-quality tables
\usepackage{amsmath}
\usepackage{amsfonts}       % blackboard math symbols
\usepackage{nicefrac}       % compact symbols for 1/2, etc.
\usepackage{microtype}      % microtypography
\usepackage{cleveref}       % smart cross-referencing
\usepackage[numbers]{natbib} % numbers for arxiv
\usepackage{doi}
\usepackage{algorithm}
\usepackage{algpseudocode}
\usepackage{fancyhdr}
\usepackage{geometry}
\usepackage{titlesec}

\hypersetup{
	pdftitle={Transformadores: Fundamentos teóricos y Aplicaciones},
	pdfsubject={CS.AI},
	pdfauthor={Jordi de la Torre},
	pdfkeywords={Aprendizaje profundo, Transformadores},
	colorlinks=true,
	linkcolor=blue,
	citecolor=blue,
	urlcolor=blue
}

\title{Transformadores: \\ Fundamentos teóricos y Aplicaciones}
\author{\href{https://orcid.org/0000-0002-8142-7983}{\includegraphics[scale=0.06]{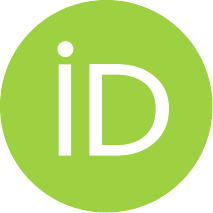}\hspace{1mm}Jordi de la Torre}\thanks{mailto:jordi.delatorre@gmail.com} \\
	Ph.D. in Computer Science and Mathematics of Security\\
	Barcelona, ES \\
	\texttt{jordi.delatorre@gmail.com}
}

\begin{document}
\maketitle
\selectlanguage{spanish} 
\begin{abstract}

Los transformadores son una arquitectura de red neuronal desarrollada originalmente para tareas de procesamiento del lenguaje natural, que se ha consolidado como la herramienta principal para abordar una amplia gama de problemas, incluyendo procesamiento de texto, audio, imágenes, aprendizaje por refuerzo, y otros dominios con datos de entrada heterogéneos. Su rasgo distintivo es el mecanismo de auto-atención, una extensión del sistema de atención propuesto en trabajos anteriores, que permite al modelo ponderar dinámicamente diferentes partes de la secuencia de entrada al procesarla. Este artículo ofrece al lector el contexto necesario para comprender investigaciones recientes sobre modelos de transformadores, y presenta los fundamentos matemáticos y algorítmicos que sustentan sus componentes principales. Además, se analizan en detalle los distintos elementos que conforman esta arquitectura, sus posibles variantes, y algunas de las aplicaciones más relevantes. El artículo se presenta en español con el objetivo de acercar este conocimiento científico a la comunidad hispanohablante.

\end{abstract}

\selectlanguage{english}
\begin{abstract}
Transformers are a neural network architecture originally developed for natural language processing, which have since become a foundational tool for solving a wide range of problems, including text, audio, image processing, reinforcement learning, and other tasks involving heterogeneous input data. Their hallmark is the self-attention mechanism, which allows the model to weigh different parts of the input sequence dynamically, and is an evolution of earlier attention-based approaches. This article provides readers with the necessary background to understand recent research on transformer models, and presents the mathematical and algorithmic foundations of their core components. It also explores the architecture's various elements, potential modifications, and some of the most relevant applications. The article is written in Spanish to help make this scientific knowledge more accessible to the Spanish-speaking community.
	
\end{abstract}

\selectlanguage{spanish}

\newpage

\section{Introducción}

Las redes neuronales completamente conectadas han sido consideradas durante décadas como aproximadores universales. En 1989, Hornik et al. demostraron que una red con una sola capa oculta y un número suficiente de nodos es capaz de aproximar cualquier función continua con la precisión deseada \cite{hornik1989multilayer}.

No obstante, a pesar de esta garantía teórica, en la práctica resulta complejo determinar tanto el número óptimo de nodos como los valores adecuados de los pesos para lograr dicha aproximación de manera eficiente.

Para hacer más tratable el proceso de optimización, se han introducido priores estructurales que restringen el espacio de búsqueda a subconjuntos más manejables y relevantes. Por ejemplo, las redes convolucionales explotan regularidades espaciales locales frecuentes en imágenes, mientras que las redes recurrentes se especializan en datos secuenciales, aprovechando patrones temporales. Estas arquitecturas están, por tanto, diseñadas para capturar regularidades estructurales del entorno, ya sean espaciales o temporales \cite{ioannou2018structural}.

Sin embargo, estas especializaciones también conllevan limitaciones. En particular, las redes recurrentes pueden tener dificultades para modelar dependencias a largo plazo debido a restricciones en su capacidad de memoria interna y a problemas de atenuación o explosión del gradiente durante el entrenamiento en secuencias extensas, lo que puede degradar su rendimiento.

Como respuesta a estas limitaciones, surgieron los modelos basados en mecanismos de atención, los cuales permiten al modelo centrarse dinámicamente en las partes más relevantes de la entrada, independientemente de su posición en la secuencia. Este enfoque mejora la capacidad para capturar dependencias de largo alcance y ha supuesto un cambio de paradigma en el diseño de arquitecturas neuronales.

\section{Evolución histórica de los mecanismos de atención}

El objetivo de esta sección no es realizar una exposición exhaustiva de todos los mecanismos de atención existentes, sino introducir el concepto y contextualizar la evolución que culminó en el sistema de auto-atención utilizado en los transformadores.

Más que un repaso detallado de cada variante, se busca presentar los fundamentos del mecanismo de atención y su progresiva sofisticación hasta convertirse en la piedra angular de la arquitectura de transformadores. Esta arquitectura ha demostrado ser altamente versátil, comparable en impacto a invenciones anteriores como las redes convolucionales o las redes recurrentes tipo LSTM. Se trata de un diseño generalista que permite integrar datos de distinta naturaleza de forma elegante y eficiente en tareas de predicción.

Uno de los hitos más relevantes en el desarrollo de redes neuronales basadas en atención ha sido la eliminación de la necesidad de recurrencia para procesar secuencias. En lugar de depender de un estado interno que se actualiza paso a paso, los transformadores permiten atender simultáneamente a todas las partes relevantes de la secuencia de entrada. Esto supone un cambio de paradigma: la dependencia temporal se transforma en una relación espacial, codificando la posición temporal como una parte adicional de la entrada.

La evolución de estas ideas comienza alrededor de 2014, impulsada por el interés en dotar a los modelos de aprendizaje profundo de cierta capacidad explicativa. Los primeros mecanismos de atención se propusieron para identificar las regiones más relevantes de una imagen en tareas de clasificación. En 2015, este enfoque se extendió al procesamiento del lenguaje natural (NLP), en particular a la traducción automática \cite{bahdanau2014neural}. Hasta entonces, las arquitecturas de codificador-decodificador utilizaban un único vector interno para representar toda la frase de entrada. Bahdanau y colaboradores propusieron un mecanismo que, en lugar de utilizar únicamente ese vector, aprovechaba también los estados intermedios del codificador, logrando mejoras significativas en rendimiento.

Este avance marcó el inicio de una intensa línea de investigación que desembocó, en 2017, en la publicación del artículo \emph{Attention is All You Need} \cite{vaswani2017attention}. En él, los autores presentaron una propuesta radical: eliminar completamente la recurrencia y diseñar una red compuesta únicamente por módulos de auto-atención, capas de normalización y capas completamente conectadas. Esta arquitectura, denominada \emph{transformer}, alcanzó resultados de nivel estado del arte en tareas de NLP, al tiempo que resolvía muchas de las dificultades inherentes al entrenamiento de redes recurrentes.

Desde entonces, los transformadores han sido objeto de una enorme expansión y refinamiento, y se han aplicado con éxito en ámbitos muy diversos: procesamiento del lenguaje natural, visión por computador, control de sistemas y más. Esta arquitectura ha revolucionado no solo el campo del lenguaje natural - tradicionalmente uno de los más complejos para el aprendizaje profundo-, sino también el procesamiento de otras señales, como las imágenes, anteriormente dominado por redes convolucionales.

\section{Retos que plantea la arquitectura de las redes neuronales recurrentes}

Las redes neuronales recurrentes (RNN, por sus siglas en inglés) surgieron con el objetivo de tratar datos secuenciales, permitiendo procesar entradas en el orden temporal en que se presentan y adaptarse a sus características dinámicas. Estas redes mantienen un estado interno que captura dependencias de corto, medio o largo alcance dentro de la secuencia, permitiendo establecer relaciones entre elementos temporales distantes con el fin de predecir salidas futuras.

Este tipo de redes se utilizan comúnmente como módulos dentro de arquitecturas más complejas para abordar diversos tipos de problemas, que pueden clasificarse en las siguientes categorías:

\begin{enumerate}
	\item \textbf{Modelos vector-secuencia:} A partir de una entrada representada por un vector de tamaño fijo, se genera una secuencia de salida de dimensión arbitraria. Un ejemplo representativo es el etiquetado automático de imágenes, donde la red genera una descripción textual a partir de una imagen.
	
	\item \textbf{Modelos secuencia-vector:} Se recibe como entrada una secuencia de longitud variable y se produce una salida de dimensión fija. Un caso típico es el análisis de sentimiento, en el que, a partir de un texto, se obtiene una etiqueta binaria que indica si el sentimiento es positivo o negativo.
	
	\item \textbf{Modelos secuencia-secuencia:} Tanto la entrada como la salida son secuencias de longitud variable. Un ejemplo clásico de esta categoría es la traducción automática, donde una frase en un idioma se traduce a otro idioma manteniendo la estructura secuencial.
\end{enumerate}

En la práctica, se ha observado que las arquitecturas basadas en redes recurrentes, diseñadas para resolver los problemas antes mencionados, no solo requieren un mayor número de épocas de entrenamiento en comparación con redes convolucionales, sino que también presentan diversas limitaciones. Entre ellas destacan: la dificultad para capturar dependencias de largo alcance debido a la profundidad de las redes, el problema de la explosión o desaparición del gradiente, y las restricciones inherentes a la paralelización del cómputo, que dificultan su escalabilidad.

Si bien arquitecturas como las LSTM (\cite{hochreiter1997long}) supusieron un avance importante al permitir el modelado efectivo de dependencias a largo plazo, estas siguen enfrentando dificultades al tratar con dependencias de muy largo alcance o secuencias extremadamente largas.

\begin{figure}[ht!]
	\centering
	\includegraphics[width=0.5\textwidth]{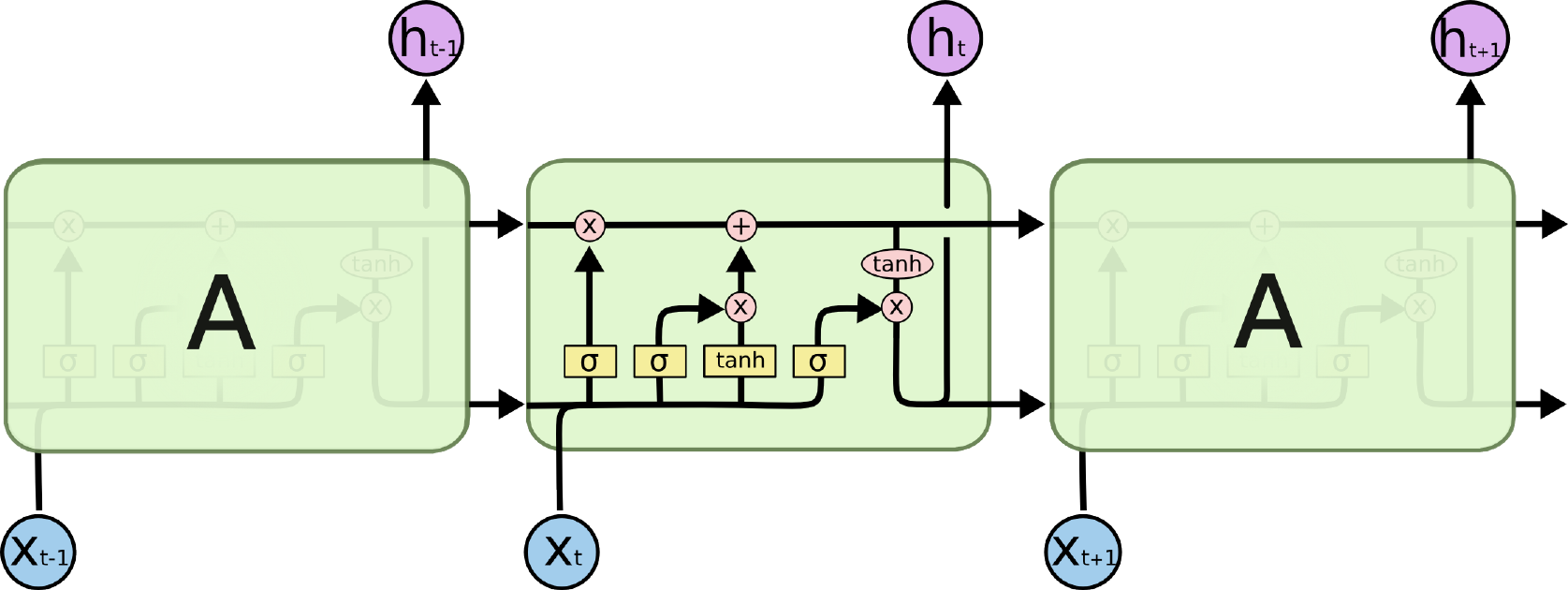}
	\label{fig:lstm}
	\caption{Diagrama básico representativo de una red recurrente con celdas LSTM}
\end{figure}

El \textit{despliegue} temporal de las redes recurrentes, necesario durante el entrenamiento, incrementa significativamente la probabilidad de aparición de los problemas de explosión y desaparición del gradiente, afectando negativamente al proceso de optimización.

Cuando las secuencias de entrada son largas, es indispensable desplegar la red en toda su extensión temporal para poder calcular adecuadamente los gradientes asociados al procesamiento secuencial. Esta operación no solo agrava los problemas mencionados relacionados con los gradientes, sino que también implica un considerable incremento en los recursos computacionales requeridos, tanto en términos de tiempo de cómputo como de uso de memoria.

\section{Arquitectura de los Transformadores}

La arquitectura de los transformadores fue diseñada para superar las limitaciones de las redes neuronales recurrentes, eliminando la recurrencia y permitiendo el procesamiento simultáneo de todos los elementos de una secuencia. Gracias al mecanismo de atención, los transformadores pueden establecer conexiones directas entre elementos distantes de la entrada, mitigando así los problemas relacionados con la desaparición o explosión del gradiente. Además, los transformadores incorporan conexiones residuales, siguiendo el enfoque propuesto por \cite{he2016deep}, lo que contribuye a facilitar el entrenamiento de redes profundas. En la práctica, el entrenamiento de los transformadores suele requerir menos épocas y, dado que permite un procesamiento completamente paralelo, resulta mucho más eficiente computacionalmente en comparación con modelos secuenciales como las RNN.

\subsection{El mecanismo de auto-atención}

El mecanismo de \textit{auto-atención} constituye el núcleo fundamental y distintivo de la arquitectura de los transformadores. Comprender en profundidad su funcionamiento resulta esencial para poder abordar las particularidades y el diseño interno de esta arquitectura.

\subsubsection{Definición formal}

Sea $\mathbf{X} \in \mathbb{R}^{d_e \times \ell_\mathcal{X}}$ una secuencia de entrada compuesta por $\ell_\mathcal{X}$ elementos, cada uno de dimensión $d_e$. De manera análoga, sea $\mathbf{Z} \in \mathbb{R}^{d_e \times \ell_\mathcal{Z}}$ una secuencia de contexto formada por $\ell_\mathcal{Z}$ elementos, también de dimensión $d_e$.

Se definen tres pares de matrices y vectores que representan transformaciones lineales aplicables a las secuencias de entrada y contexto: 
\[
(\mathbf{W_q}, \mathbf{b_q}) \in \mathbb{R}^{d_{\text{attn}} \times d_e} \times \mathbb{R}^{d_{\text{attn}}}, \quad
(\mathbf{W_k}, \mathbf{b_k}) \in \mathbb{R}^{d_{\text{attn}} \times d_e} \times \mathbb{R}^{d_{\text{attn}}}, \quad
(\mathbf{W_v}, \mathbf{b_v}) \in \mathbb{R}^{d_{\text{out}} \times d_e} \times \mathbb{R}^{d_{\text{out}}}.
\]

A partir de estas transformaciones se obtienen las matrices de consultas, claves y valores, respectivamente:
\[
\mathbf{Q} = \mathbf{W_q} \mathbf{X} + \mathbf{b_q} \mathbf{1}^\top \in \mathbb{R}^{d_{\text{attn}} \times \ell_\mathcal{X}}, \quad
\mathbf{K} = \mathbf{W_k} \mathbf{Z} + \mathbf{b_k} \mathbf{1}^\top \in \mathbb{R}^{d_{\text{attn}} \times \ell_\mathcal{Z}}, \quad
\mathbf{V} = \mathbf{W_v} \mathbf{Z} + \mathbf{b_v} \mathbf{1}^\top \in \mathbb{R}^{d_{\text{out}} \times \ell_\mathcal{Z}}.
\]

La matriz de similitud $\mathbf{S}$ se define como:
\[
\mathbf{S} = \frac{1}{\sqrt{d_{\text{attn}}}} \mathbf{K}^\top \mathbf{Q} \in \mathbb{R}^{\ell_\mathcal{Z} \times \ell_\mathcal{X}},
\]
la cual mide la compatibilidad entre cada elemento de la secuencia de entrada y los elementos de la secuencia de contexto.

La ecuación de atención que define la interacción de la secuencia de entrada $\mathbf{X}$ con la secuencia de contexto $\mathbf{Z}$ se expresa de la siguiente manera:

\begin{equation}
	\tilde{\mathbf{V}} = \mathbf{V} \cdot \text{softmax}(\mathbf{S}),
	\quad \text{donde} \quad \tilde{\mathbf{V}} \in \mathbb{R}^{d_{\text{out}} \times \ell_\mathcal{X}}.
	\label{eq:self-attention}
\end{equation}

Aquí, $\tilde{\mathbf{V}}$ representa la salida ponderada de los valores $\mathbf{V}$ tras aplicar la normalización de similitud $\mathbf{S}$ a través de la función \textit{softmax}.

La función $\text{softmax}$ se aplica de forma independiente a cada fila de la matriz de similitudes $\mathbf{S}$, de modo que cada elemento de la matriz depende de todos los demás elementos dentro de la misma fila.

Cuando la secuencia de entrada y la secuencia de contexto coinciden, es decir, cuando $\mathbf{X} = \mathbf{Z}$, hablamos de atención a la propia secuencia, o más brevemente, \textit{auto-atención}.

El algoritmo \ref{alg:position-single-attention} presenta la versión algorítmica que describe el cálculo de la auto-atención.

\begin{algorithm}
	\caption{Esquema básico de atención con una única consulta (versión secuencial)}
	\label{alg:position-single-attention}
	\begin{algorithmic}[1]
		\State \textbf{Entrada:} $\mathbf{e} \in \mathbb{R}^{d_{\text{in}}}$, la representación del token de entrada actual
		\State \textbf{Entrada:} $\mathbf{e}_t \in \mathbb{R}^{d_{\text{in}}}$, representación vectorial de los tokens de contexto $t \in [T]$
		\State \textbf{Salida:} $\tilde{\mathbf{v}} \in \mathbb{R}^{d_{\text{out}}}$, representación vectorial combinada del token y contexto.
		\State \textbf{Parámetros:} $\mathbf{W}_q, \mathbf{W}_k \in \mathbb{R}^{d_{\text{attn}} \times d_{\text{in}}}$, parámetros de proyección lineal de consulta y clave.
		\State \textbf{Parámetros:} $\mathbf{b}_q, \mathbf{b}_k \in \mathbb{R}^{d_{\text{attn}}}$, sesgos de proyección de consulta y clave.
		\State \textbf{Parámetros:} $\mathbf{W}_v \in \mathbb{R}^{d_{\text{out}} \times d_{\text{in}}}$, parámetro de proyección lineal de los valores.
		\State \textbf{Parámetros:} $\mathbf{b}_v \in \mathbb{R}^{d_{\text{out}}}$, sesgo de proyección de los valores.
		\State
		\State \textbf{Algoritmo:}
		\begin{algorithmic}[1]
			\State $\mathbf{q} \leftarrow \mathbf{W}_q \mathbf{e} + \mathbf{b}_q$
			\State $\forall t \in [T], \ \mathbf{k}_t \leftarrow \mathbf{W}_k \mathbf{e}_t + \mathbf{b}_k$
			\State $\forall t \in [T], \ \mathbf{v}_t \leftarrow \mathbf{W}_v \mathbf{e}_t + \mathbf{b}_v$
			\State $\forall t \in [T], \ \alpha_t \leftarrow \frac{\exp\left(\mathbf{q}^\top \mathbf{k}_t / \sqrt{d_{\text{attn}}}\right)}{\sum_{u=1}^{T} \exp\left(\mathbf{q}^\top \mathbf{k}_u / \sqrt{d_{\text{attn}}}\right)}$
			\State \textbf{return} $\tilde{\mathbf{v}} = \sum_{t=1}^{T} \alpha_t \mathbf{v}_t$
		\end{algorithmic}
	\end{algorithmic}
\end{algorithm}

\begin{figure}[ht!]
	\centering
	\includegraphics[width=0.3\textwidth]{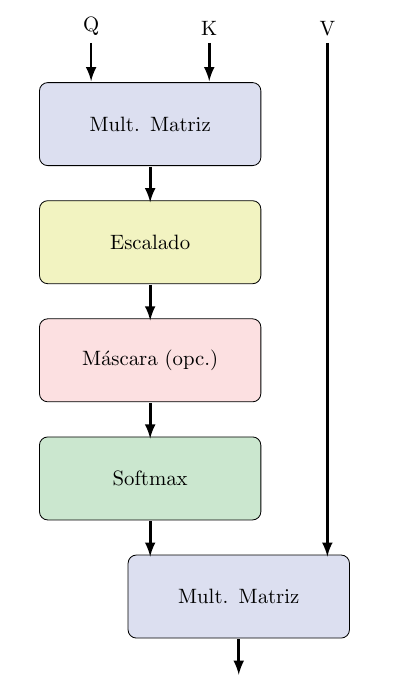}
	\label{fig:atencion}
	\caption{Diagrama de bloques representativo de las operaciones matriciales involucradas en los cálculos de la atención}	
\end{figure}

Como se indica en la versión algorítmica presentada en \ref{alg:position-single-attention}, este cálculo se realiza para todos los elementos de la secuencia. Con el objetivo de optimizar el tiempo de computación, es de interés paralelizar el cálculo en la medida de lo posible. Una forma de lograrlo es mediante la configuración matricial del algoritmo. En la Figura~\ref{fig:atencion} se presenta, mediante un diagrama de bloques, las operaciones matriciales necesarias para llevar a cabo el proceso. El módulo \textit{Máscara}, que aparece en el diagrama y está marcado como opcional, es una operación que, en caso de ser necesario, permite anular el efecto de parte de las secuencias.

El algoritmo \ref{alg:position-masked-attention} es la versión matricial alternativa del algoritmo \ref{alg:position-single-attention} que explota las posibilidades de cálculo en paralelo, acelerando así el proceso de computación.

La operación de enmascarado permite diseñar estrategias de entrenamiento alternativas para abordar distintos problemas utilizando la misma arquitectura. El sistema de \textit{auto-atención sin máscara} trata toda la secuencia como contexto, es decir, $\mathbf{Z} = \mathbf{X}$, con $(\text{Mask} \equiv 1)$, lo que da lugar a transformadores con contextos bidireccionales. Al aplicar máscaras a los elementos de la secuencia situados en uno de los lados, obtenemos transformadores de contexto unidireccional. Además, cuando se tienen dos secuencias de entrada, es posible aplicar la secuencia cruzada, enmascarando los elementos de una de las secuencias de entrada y utilizando la otra como contexto.

\begin{algorithm}[H]
	\caption{$\hat{\mathbf{V}} \leftarrow \text{atencion}(\mathbf{X}, \mathbf{Z} \mid \mathbf{W}_{qkv}, \text{Mask})$ (versión matricial)}
	\begin{algorithmic}[1]
		\State \textbf{Entrada:} $\mathbf{X} \in \mathbb{R}^{d_X \times \ell_X}$ y $\mathbf{Z} \in \mathbb{R}^{d_Z \times \ell_Z}$, representaciones vectoriales de las secuencias principal y de contexto.
		\State \textbf{Salida:} $\hat{\mathbf{V}} \in \mathbb{R}^{d_{\text{out}} \times \ell_X}$, representación vectorial actualizada de $\mathbf{X}$, recogiendo la información de la secuencia $\mathbf{Z}$.
		\State \textbf{Parámetros:} $\mathbf{W}_{qkv}$, consistente en:
		\begin{itemize}
			\item $\mathbf{W}_q \in \mathbb{R}^{d_{\text{attn}} \times d_X}$, $\mathbf{b}_q \in \mathbb{R}^{d_{\text{attn}}}$: parámetros de proyección para la consulta.
			\item $\mathbf{W}_k \in \mathbb{R}^{d_{\text{attn}} \times d_Z}$, $\mathbf{b}_k \in \mathbb{R}^{d_{\text{attn}}}$: parámetros de proyección para la clave.
			\item $\mathbf{W}_v \in \mathbb{R}^{d_{\text{out}} \times d_Z}$, $\mathbf{b}_v \in \mathbb{R}^{d_{\text{out}}}$: parámetros de proyección para el valor.
		\end{itemize}
		\State \textbf{Hiperparámetros:} $Mask \in \{0,1\}^{\ell_Z \times \ell_X}$, máscara que especifica qué elementos de la secuencia deben ser enmascarados.
		\State
		\begin{tabular}{ l l r }
			\textbf{1)} & $\mathbf{Q} \leftarrow \mathbf{W}_q \mathbf{X} + \mathbf{b}_q \mathbf{1}^T$ & [[\textbf{Q}uery $\in \mathbb{R}^{d_{\text{attn}} \times \ell_X}$]] \\
			\textbf{2)} & $\mathbf{K} \leftarrow \mathbf{W}_k \mathbf{Z} + \mathbf{b}_k \mathbf{1}^T$ & [[\textbf{K}ey $\in \mathbb{R}^{d_{\text{attn}} \times \ell_Z}$]] \\
			\textbf{3)} & $\mathbf{V} \leftarrow \mathbf{W}_v \mathbf{Z} + \mathbf{b}_v \mathbf{1}^T$ & [[\textbf{V}alue $\in \mathbb{R}^{d_{\text{out}} \times \ell_Z}$]] \\
			\textbf{4)} & $\mathbf{S} \leftarrow \mathbf{K}^T \mathbf{Q}$ & [[\textbf{S}core $\in \mathbb{R}^{\ell_Z \times \ell_X}$]] \\
			\textbf{5)} & $\forall t_z, t_x, \text{ if } \neg Mask[t_z, t_x] \text{ then } S[t_z,t_x] \leftarrow -\infty$ & [[Aplica la máscara]] \\
			\textbf{6)} & \textbf{return} $\hat{\mathbf{V}} = \mathbf{V} \cdot \text{softmax}\left(\frac{\mathbf{S}}{\sqrt{d_{\text{attn}}}}\right)$ & [[Salida de la atención]] \\
		\end{tabular}
	\end{algorithmic}
	\label{alg:position-masked-attention}
\end{algorithm}

Informalmente, el transformador utiliza una serie de características de la secuencia de entrada para realizar una combinación ponderada de ellas, basándose en su similitud. Los pesos asignados a cada característica se calculan a partir de la similitud entre los pares de características de entrada. Este proceso se repite a lo largo de las capas de la red. En la primera capa, se comparan los pares de características, y en las capas posteriores se comparan pares de pares, y así sucesivamente. A medida que se profundiza en la red, el número de características combinadas aumenta exponencialmente, lo que permite obtener representaciones cada vez más complejas y variadas de la secuencia, a medida que se llega a las capas finales.

\subsubsection{Analogía con las bases de datos}

Los nombres originales utilizados para definir las transformaciones lineales empleadas en el cálculo de la auto-atención hacen referencia a una analogía conceptual con las consultas en bases de datos relacionales.

En una base de datos, tenemos consultas (\emph{queries}, $\mathbf{Q}$), claves (\emph{keys}, $\mathbf{K}$) y valores (\emph{values}, $\mathbf{V}$). Al realizar una consulta $\mathbf{Q}$ sobre un conjunto de claves $\mathbf{K}_1, \mathbf{K}_2, \dots, \mathbf{K}_N$, la base de datos reporta como resultado una serie de valores $\mathbf{V}_1, \mathbf{V}_2, \dots, \mathbf{V}_N$.

El mecanismo de atención de los transformadores puede considerarse una versión probabilística de este proceso. Una función de similaridad compara la consulta $\mathbf{Q}$ con cada una de las claves $\mathbf{K}$. El resultado es un vector que puede interpretarse como la medida de similaridad entre la consulta $\mathbf{Q}$ y cada una de las claves $\mathbf{K}$. Este valor se utiliza posteriormente para calcular un peso que sirve para obtener el valor final, como una combinación lineal de los valores de entrada. La función de similaridad puede definirse de distintas maneras (ver ecuación \ref{eq:similaridad}). La elección de una u otra depende de la decisión de diseño. Como hemos observado en las ecuaciones y algoritmos presentados en el apartado anterior, los transformadores que estudiaremos utilizan como función de similaridad el producto escalar escalado.

\begin{equation}
	s_i = f(\mathbf{Q}, \mathbf{K}_i) =
	\begin{cases}
		\mathbf{Q}^T \mathbf{K}_i & \text{producto escalar} \\
		\frac{\mathbf{Q}^T \mathbf{K}_i}{\sqrt{d}} & \text{producto escalar escalado} \\
		\mathbf{Q}^T \mathbf{W} \mathbf{K}_i & \text{producto escalar general} \\
		\mathbf{W}_Q^T + \mathbf{W}_K^T \mathbf{K}_i & \text{similaridad aditiva} \\
		\text{otros} & \text{kernels, etc.}
	\end{cases}
	\label{eq:similaridad}
\end{equation}

Una vez calculada la similaridad, obtenemos, para una consulta concreta \(\mathbf{Q}\), un valor por cada clave, es decir, un vector de dimensión igual al número de claves. Los valores altos de similaridad indican un alto grado de coincidencia. Por lo tanto, dado que la intención es crear una función de búsqueda probabilística, se utiliza ese valor para calcular la probabilidad de coincidencia aplicando una función \textit{softmax} (ver ecuación \ref{eq:probabilidad}), obteniendo como resultado la probabilidad de la posición \(i\) para la pareja consulta-clave determinada.

Finalmente, se obtiene un conjunto de valores como combinación lineal de los valores fuente, ponderados por el peso de cada uno de ellos en la comparación entre consulta y clave (ver ecuación \ref{eq:atencion}).

\begin{equation}
	\omega_i = \frac{\exp{(s_i)}}{\sum_j \exp(s_j)} 
	\label{eq:probabilidad}
\end{equation}

\begin{equation}
	A = \sum_i \omega_i V_i 
	\label{eq:atencion}
\end{equation}

\subsubsection{Auto-atención múltiple}

El mecanismo descrito en las secciones anteriores permite realizar una transformación utilizando el súper-conjunto \((\boldsymbol{Q}, \boldsymbol{K}, \boldsymbol{V})\) desde el espacio inicial \(\mathbb{R}^{n \times d}\), lo que da como resultado un valor en el espacio \(\mathbb{R}^{n \times d_v}\).

Una forma de ampliar las capacidades del sistema es aplicar varias transformaciones en paralelo del mismo tipo, pero utilizando diferentes conjuntos \((\boldsymbol{Q_i}, \boldsymbol{K_i}, \boldsymbol{V_i})\) para cada cabeza de atención. Cada uno de estos conjuntos devuelve un valor distinto, ubicado en el espacio \(\mathbb{R}^{n \times d_v}\). Esto se conoce como **atención múltiple** o **multi-head attention**.

La ecuación que se utiliza para calcular la atención múltiple es la siguiente:

\[
\boldsymbol{M} = \text{Concat}_{i=1}^{h} \left[ \boldsymbol{D}_i \left( \boldsymbol{Q}_i, \boldsymbol{K}_i, \boldsymbol{V}_i \right) \right] \boldsymbol{W}_O
\]

Donde:
- \( \boldsymbol{Q}_i, \boldsymbol{K}_i, \boldsymbol{V}_i \) corresponden a los conjuntos de consultas (queries), claves (keys) y valores (values) para la \(i\)-ésima cabeza de atención.
- \( \boldsymbol{D}_i \left( \boldsymbol{Q}_i, \boldsymbol{K}_i, \boldsymbol{V}_i \right) \) es el resultado de aplicar la operación de atención sobre cada conjunto \( \left( \boldsymbol{Q}_i, \boldsymbol{K}_i, \boldsymbol{V}_i \right) \), dando como resultado una matriz de dimensión \( \mathbb{R}^{n \times d_v} \).
- \( \text{Concat}_{i=1}^{h} \) es la operación de concatenación de los resultados de las \(h\) cabezas de atención, lo que produce una matriz de dimensión \( \mathbb{R}^{n \times h \cdot d_v} \).
- \( \boldsymbol{W}_O \) es una matriz de pesos de dimensión \( \mathbb{R}^{h \cdot d_v \times d_{\text{out}}} \), que se aplica para proyectar el resultado concatenado a la dimensión final de salida \( d_{\text{out}} \).

Este mecanismo de atención múltiple permite que el modelo capture diferentes aspectos de la información en paralelo, mejorando su capacidad para aprender representaciones complejas de las entradas.

Si el número de bloques de atención paralelos \( n \) se escoge de forma tal que \( n = \frac{d}{d_v} \), entonces las dimensiones del espacio de salida serán iguales a las del espacio de entrada, es decir, \( \mathbb{R}^{n \times d} \). La matriz \( \boldsymbol{W}_O \) realiza una transformación lineal posterior a la concatenación y permitiría eventualmente la modificación de las dimensiones del espacio de salida, en caso de que en el diseño se desee modificar dicha dimensión.

La utilización del mecanismo de atención múltiple permite aplicar simultáneamente diferentes transformaciones a los pares de atributos de entrada, lo que incrementa tanto la diversidad como la complejidad de las comparaciones. Además, posibilita mantener la dimensión del espacio de trabajo, lo cual puede resultar útil en arquitecturas que emplean bloques de la misma naturaleza apilados unos sobre otros.

El algoritmo \ref{alg:multi-head-attention} muestra la versión algorítmica del proceso.

\begin{algorithm}[H]
	\caption{$\hat{\boldsymbol{V}} \leftarrow \text{atencion\_multicabezal}(\boldsymbol{X}, \boldsymbol{Z} \mid \boldsymbol{W_{qkv}}, \text{Mask})$ (versión matricial)} 
	\begin{algorithmic}
		\State /* Calcula la atención multi-cabezal (con máscara) */
		\State \textbf{Entrada:} $\boldsymbol{X} \in \mathbb{R}^{d_X \times \ell_X}$ y $\boldsymbol{Z} \in \mathbb{R}^{d_Z \times \ell_Z}$, representaciones vectoriales de las secuencias principal y de contexto.
		\State \textbf{Salida:} $\tilde{\boldsymbol{V}} \in \mathbb{R}^{d_{\text{out}} \times \ell_X}$, representación vectorial actualizada de $\boldsymbol{X}$, que recoge la información de la secuencia $\boldsymbol{Z}$.
		\State \textbf{Parámetros:} $\boldsymbol{W_{qkv}}$, consistente en:
		\begin{itemize}
			\item $\boldsymbol{W_q} \in \mathbb{R}^{d_{\text{attn}} \times d_X}$, $\boldsymbol{b_q} \in \mathbb{R}^{d_{\text{attn}}}$,
			\item $\boldsymbol{W_k} \in \mathbb{R}^{d_{\text{attn}} \times d_Z}$, $\boldsymbol{b_k} \in \mathbb{R}^{d_{\text{attn}}}$,
			\item $\boldsymbol{W_v} \in \mathbb{R}^{d_{\text{out}} \times d_Z}$, $\boldsymbol{b_v} \in \mathbb{R}^{d_{\text{out}}}$.
		\end{itemize}
		\State \textbf{Hiperparámetros:} $Mask \in \{0,1\}^{\ell_Z \times \ell_X}$.
		\State
		\begin{tabular}{ l l }
			\textbf{1)} & Para $h \in [H]$: \\
			\textbf{2)} & $\boldsymbol{Y^h} \leftarrow \text{atencion}(\boldsymbol{X}, \boldsymbol{Z} \mid \boldsymbol{W_{qkv}^h}, Mask)$ \\
			\textbf{3)} & $\boldsymbol{Y} \leftarrow [\boldsymbol{Y^1}; \boldsymbol{Y^2}; \dots; \boldsymbol{Y^H}]$ \\
			\textbf{4)} & return $\hat{\boldsymbol{V}} = \boldsymbol{W_o} \boldsymbol{Y} + \boldsymbol{b_o} \boldsymbol{1}^T$ \\
		\end{tabular}
	\end{algorithmic}
	\label{alg:multi-head-attention}
\end{algorithm}

\subsection{Transformadores}

Llegados a este punto, estamos en condiciones de presentar la arquitectura del \textit{transformador}. Como ya hemos indicado anteriormente, este nuevo paradigma de diseño de red neuronal fue introducido con la publicación de \cite{vaswani2017attention} y se planteó inicialmente para su aplicación en sistemas de traducción automática.

\begin{figure}[ht!]
	\centering
	\includegraphics[width=0.6\textwidth]{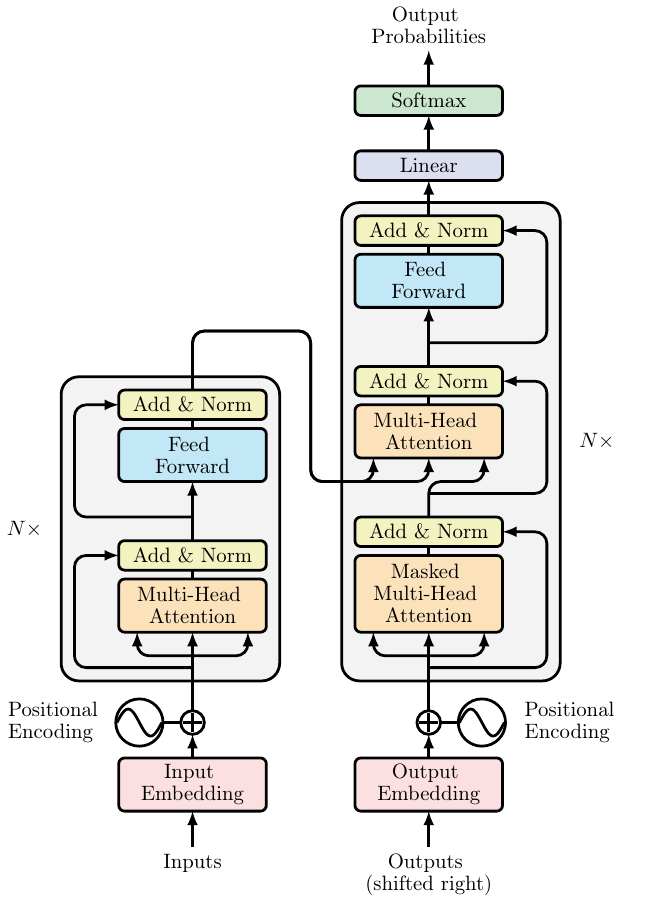}
	\caption{Arquitectura del transformador introducida en ``Attention is all you need'' \cite{vaswani2017attention}}
	\label{fig:transformador}
\end{figure}

En la Figura~\ref{fig:transformador} se muestra el diagrama de los bloques integrantes de la red de transformadores tal como fue presentado por primera vez en \cite{vaswani2017attention}. Cada elemento de la secuencia de entrada se convierte en una representación numérica (\textit{input embedding}), que posteriormente se combina (mediante suma en este caso) con una función que codifica la posición del elemento en la secuencia (\textit{positional encoding}). Este valor sirve como entrada a una pila de \(N\) bloques denominados \textit{codificadores}. Los atributos obtenidos tras la codificación son uno de los conjuntos de entradas que alimentarán a la red de \textit{decodificadores}. 

El objetivo de esta red es predecir la siguiente palabra, utilizando no solo la información proveniente de la cadena de entrada a través del codificador, sino también todos los elementos de la cadena de salida producidos hasta el momento.

En las secciones siguientes, describiremos con detalle cada una de las partes integrantes de la arquitectura, agrupándolas en sus elementos constitutivos de mayor nivel.

\subsubsection{Notación}

Sea \( V \) un conjunto finito denominado \textit{vocabulario}, denotado como \( [N_V] := \lbrace 1, \dots, N_V \rbrace \). Este conjunto puede estar formado por letras o palabras completas, aunque típicamente está compuesto por partes constituyentes de palabras denominadas \textit{tokens}.

Sea \( \mathbf{x} \equiv x[1 : \ell] \equiv x[1] x[2] \dots x[\ell] \in V^* \) una secuencia de \textit{tokens}, por ejemplo, una frase, un párrafo o un documento.

Siguiendo la notación matemática y en contra de lo establecido en algunos lenguajes de programación como C o Python, el primer índice de matrices y vectores utilizados en este módulo será el uno. Es decir, \( X[1 : \ell] \), por ejemplo, se refiere a la cadena de caracteres que va desde el primer hasta el elemento \( \ell \), ambos incluidos.

Para una matriz \( M \in \mathbb{R}^{d \times d'} \), escribimos \( M[i, :] \in \mathbb{R}^{d'} \) para referirnos a la fila \( i \) y \( M[:, j] \in \mathbb{R}^{d} \) para la columna \( j \).

Utilizamos la convención matriz \( \times \) columna más común en matemáticas, en lugar de fila \( \times \) matriz, que es más típica en la mayoría de la literatura de transformadores. Es decir, las matrices están traspuestas.

\subsubsection{Tratamiento de entrada}

\begin{figure}[ht!]
	\centering
	\includegraphics[width=0.3\textwidth]{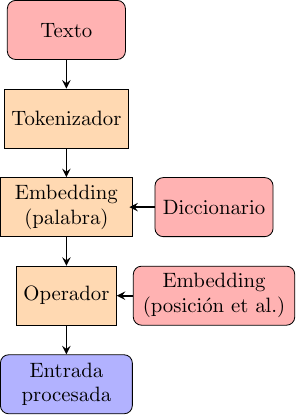}
	\label{fig:capa-entrada}
	\caption{Esquema típico de las operaciones necesarias para convertir una entrada de texto a formato necesario para ser tratado por el transformador}
\end{figure}

En la Figura~\ref{fig:capa-entrada} se muestra el proceso de tratamiento de entrada propio del transformador. El texto se alimenta a un \textit{tokenizador}, cuya función es particionar la secuencia en sus elementos constituyentes. Cada elemento es representado por un vector ortogonal que tiene un \( 1 \) en la posición correspondiente al diccionario identificativo del elemento y ceros en el resto. A continuación, se realiza una transformación lineal para comprimir la representación inicial en un vector denso de dimensión menor. Finalmente, mediante un operador (que suele ser la suma, pero nada impide que pudiera ser, por ejemplo, la concatenación u otro distinto), se añade la información referente a la posición que ocupa el elemento dentro de la secuencia. La representación numérica de la secuencia ya está lista para ser alimentada a los bloques de procesamiento siguientes.

La capa de entrada se encarga del preprocesamiento de los datos para adecuarlos a un formato que sea interpretable por el transformador. Los transformadores se diseñaron inicialmente como una forma alternativa de tratamiento de los datos secuenciales. El primer paso para el tratamiento de este tipo de datos es la partición de la secuencia en sus elementos constitutivos. Este proceso es conocido habitualmente como \textit{tokenización}. Esta partición se puede hacer de muchos modos: a nivel de palabra, partes de palabra o bien caracteres. 

En los diseños actuales de transformadores, lo más habitual es utilizar particiones a nivel de partes de palabra. Muchas palabras están constituidas por distintos monemas (unidades mínimas de significado). Particionar a nivel de palabra tiene el inconveniente de que muchas de las unidades elementales de significado no se separan, lo que suele dificultar el funcionamiento de modelos interpretativos del lenguaje. 

De entrada, se podría conjeturar que una separación a nivel de carácter permitiría al modelo reconstruir las unidades mínimas de significado. Sin embargo, los artículos publicados indican explícitamente que los resultados prácticos, al menos hasta el momento, muestran que la partición a nivel de carácter es demasiado agresiva, con resultados significativamente inferiores a los obtenidos utilizando particiones a nivel de partes de palabra. 

Entre los particionadores más típicos utilizados con transformadores tenemos \textit{BPE} (\textit{Byte Pair Encoding}) \cite{gage1994new-BPE}, \textit{Unigram} \cite{kudo2018subword-unigram} y \textit{WordPiece} \cite{devlin2018bert}. La diferencia entre estos radica en la estrategia seguida para escoger los pares de caracteres a concatenar, con el fin de elegir cada una de las subpalabras para formar el diccionario. 

Cada modelo preentrenado está diseñado para ser usado con un diccionario específico derivado del particionador escogido.

Para el caso de los transformadores aplicados al tratamiento de imagen, la propuesta existente hasta el momento consiste en particionar la imagen en trozos no superpuestos de dimensiones reducidas y alimentarlos secuencialmente. 

Si bien la preparación de las entradas descritas es el modo actual establecido de hacerlo, no existe una razón para que esto deba ser de otro modo. Por lo tanto, es posible que en el futuro surjan maneras alternativas de codificar las entradas.

\begin{algorithm}[H]
	\caption{Embedding de un token}
	\begin{algorithmic}[1]
		\State \textbf{Entrada:} $v \in V \cong [N_V]$, identificador del token (p.ej. one-hot)
		\State \textbf{Salida:} $\mathbf{e} \in \mathbb{R}^{d_e}$, la representación vectorial del token
		\State \textbf{Parámetros:} $\mathbf{W}_e \in \mathbb{R}^{d_e \times N_V}$, matriz de embedding
		\State \textbf{Operación:} $\mathbf{e} = \mathbf{W}_e[:, v]$
		\State \textbf{Retorno:} $\mathbf{e}$
	\end{algorithmic}
	\label{alg:token-embedding}
\end{algorithm}

\begin{algorithm}[H]
	\caption{Embedding de la posición}
	\begin{algorithmic}[1]
		\State \textbf{Entrada:} $\ell \in \ell_{\text{max}}$, posición del token dentro de la secuencia
		\State \textbf{Salida:} $\mathbf{e_p} \in \mathbb{R}^{d_e}$, la representación vectorial de la posición
		\State \textbf{Parámetros:} $\mathbf{W_p} \in \mathbb{R}^{d_e \times \ell_{\text{max}}}$, matriz de embeddings de posición
		\State \textbf{Operación:} $\mathbf{e_p} = \mathbf{W_p}[:, \ell]$
		\State \textbf{Retorno:} $\mathbf{e_p}$
	\end{algorithmic}
	\label{alg:position-embedding}
\end{algorithm}

\subsubsection{Codificadores}

El codificador está compuesto por un cabezal de auto-atención múltiple seguido de una red completamente conectada. Ambas capas forman parte de una arquitectura residual, lo que implica que, después de cada una de estas capas, se realiza una suma de la entrada original con la salida de la capa correspondiente, seguida de una normalización. 

\begin{figure}[ht!]
	\centering
	\includegraphics[width=0.3\textwidth]{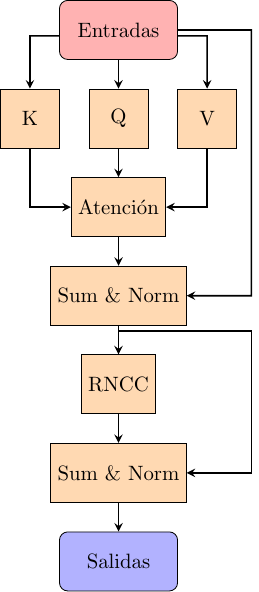}
	\label{fig:codificador}
	\caption{Esquema típico de los elementos integrantes de un codificador. La red de codificación integra varios de estos elementos en serie. La salida de uno sirve de entrada para el siguiente}
\end{figure}

En la Figura~\ref{fig:codificador} se presenta el diagrama de bloques que ilustra los componentes que conforman el codificador.

\subsubsection{Decodificadores}

El decodificador consta de un bloque principal que incluye un cabezal de auto-atención múltiple, el cual recibe como entradas los elementos de la cadena de salida de la red del transformador desplazados una posición. Para garantizar que la predicción dependa únicamente de la información precedente y no de la posterior, se incorpora una matriz de enmascarado que anula el efecto de los elementos posteriores en la cadena de salida.

A continuación, se encuentra un segundo bloque que recibe como entrada tanto la salida del primer bloque como la salida proveniente de la codificación de la cadena de entrada. La salida de este bloque se alimenta a una capa completamente conectada. Todas estas etapas siguen una estructura residual, y cada una está acompañada de una suma con la entrada original y su correspondiente normalización. La red de decodificación apila en serie $N$ elementos de este tipo.

\begin{figure}[ht!]
	\centering
	\includegraphics[width=0.4\textwidth]{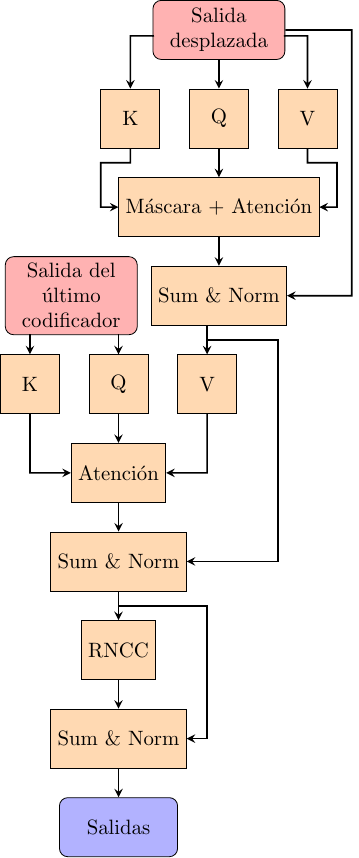}
	\label{fig:decodificador}
	\caption{Esquema típico de los elementos integrantes de un decodificador. La red de decodificación integra varios de estos elementos en serie. La salida de uno sirve de entrada para el siguiente. A todos ellos se alimenta la salida de la red de codificadores}
\end{figure}

En la Figura~\ref{fig:decodificador} se muestra el diagrama de bloques que ilustra los componentes del decodificador.

\subsubsection{Tratamiento de salida}

\begin{figure}[ht!]
	\centering
	\includegraphics[width=0.18\textwidth]{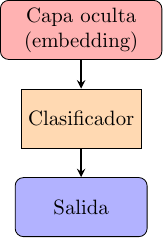}
	\label{fig:capa-salida}
	\caption{Esquema típico de las operaciones necesarias para convertir la última capa oculta del transformador en una salida}
\end{figure}

En la capa de salida, es necesario realizar una conversión desde el espacio de representación interna al espacio de representación externa (ver Figura~\ref{fig:capa-salida}). Esta transformación es inversa a la operación de \textit{embedding}, donde se pasa de un vector representativo del diccionario a una representación interna.

Existen dos enfoques para llevar a cabo la conversión: el primero consiste en tratar la capa de salida como independiente, aprendiendo los pesos de conversión de manera separada; el segundo consiste en utilizar la matriz inversa del \textit{embedding}, sin realizar un aprendizaje independiente.

En \cite{vaswani2017attention} se adopta la estrategia definida en \cite{press2016using} de vincular la transformación de la capa de entrada con la de la capa de salida. Uno de los pasos iniciales para preparar las entradas consiste en convertir la representación \textit{one-hot} de la posición de un token en el diccionario a una representación "densa" con las dimensiones del modelo. Este paso se lleva a cabo mediante una transformación lineal que reduce las dimensiones desde las del diccionario hasta las dimensiones internas del modelo. La idea presentada en \cite{press2016using} y posteriormente utilizada en \cite{vaswani2017attention} consiste en emplear la misma matriz de transformación tanto para las entradas como para las salidas. De esta manera, el clasificador de salida se convierte en una transformación lineal con los mismos pesos que la transformación de entrada. 

El algoritmo \ref{alg:unembedding} muestra la versión algorítmica de este proceso.

\begin{algorithm}[H]
	\caption{Embedding Inverso} 
	\begin{algorithmic}[1]
		\State /* Convierte la representación interna de la red en un elemento */
		\State \textbf{Entrada:} $\boldsymbol{e} \in \mathbb{R}^{d_e}$, representación interna de un elemento
		\State \textbf{Salida:} $\boldsymbol{p} \in \Delta(V)$, vector que representa la distribución de probabilidad sobre el conjunto de elementos del vocabulario
		\State \textbf{Parámetros:} $\boldsymbol{W_u} \in \mathbb{R}^{N_V \times d_e}$, matriz de embedding inversa
		\State \textit{Operación:} $\boldsymbol{p} = \text{softmax}(\boldsymbol{W_u} \boldsymbol{e})$
		\State \textbf{Retorno:} $\boldsymbol{p}$
	\end{algorithmic}
	\label{alg:unembedding}
\end{algorithm}

\subsubsection{Normalización de capas}

Las capas de normalización están presentes como un bloque constituyente tanto en los codificadores como en los decodificadores. Su propósito es controlar explícitamente la media y la varianza de la activación de cada neurona.

El algoritmo \ref{alg:layer-normalization} describe detalladamente el proceso que ocurre en estas capas.

\begin{algorithm}[H]
	\caption{$\hat{\boldsymbol{e}} \leftarrow \text{normalizacion\_de\_capa}(\boldsymbol{e} | \boldsymbol{\gamma}, \boldsymbol{\beta})$} 
	\begin{algorithmic}[1]
		\State /* Normaliza las activaciones $\boldsymbol{e}$ de la capa */
		\State \textbf{Entrada:} $\boldsymbol{e} \in \mathbb{R}^{d_e}$, activaciones de entrada sin normalizar
		\State \textbf{Salida:} $\hat{\boldsymbol{e}} \in \mathbb{R}^{d_e}$, activaciones normalizadas
		\State \textbf{Parámetros:} $\boldsymbol{\gamma}, \boldsymbol{\beta} \in \mathbb{R}^{d_e}$, vector de escalado y desplazamiento
		\State
		\begin{tabular}{ l l }
			\textbf{1)} & $\boldsymbol{m} \leftarrow \frac{1}{d_e} \sum_{i=1}^{d_e} \boldsymbol{e}[i]$ \\
			\textbf{2)} & $v \leftarrow \frac{1}{d_e} \sum_{i=1}^{d_e} (\boldsymbol{e}[i] - \boldsymbol{m})^2$ \\
			\textbf{3)} & \textbf{return} $\hat{\boldsymbol{e}} = \frac{\boldsymbol{e} - \boldsymbol{m}}{\sqrt{v}} \odot \boldsymbol{\gamma} + \boldsymbol{\beta}$, donde $\odot$ denota la multiplicación componente a componente
		\end{tabular}
	\end{algorithmic}
	\label{alg:layer-normalization}
\end{algorithm}

\section{Tipos de Transformadores}

En la sección anterior se ha presentado la arquitectura del transformador, mostrando el diagrama original introducido en \cite{vaswani2017attention}. Como ya se indicó previamente, la propuesta inicial surgió como una solución alternativa para problemas de traducción de secuencia a secuencia y representa el primer modelo de transformador en su tipo. En este apartado, se discutirán las modificaciones de la arquitectura original que se desarrollaron posteriormente, las cuales dieron lugar a subfamilias especializadas en la resolución de distintos problemas. Actualmente, se pueden distinguir tres tipos principales de modelos o arquitecturas: los auto-regresivos, los de auto-codificación y los de conversión de secuencia a secuencia.

\subsection{Transformadores secuencia-a-secuencia}

Los modelos de secuencia-a-secuencia utilizan dos unidades básicas de diseño denominadas codificadores y decodificadores, cuyo objetivo es la conversión de una secuencia de entrada en otra distinta. Sus aplicaciones principales incluyen la traducción, el resumen y la respuesta a preguntas. El modelo original es un ejemplo típico de este tipo de aplicaciones. \textbf{T5} \cite{raffel2020exploring} es un ejemplo que fue presentado posteriormente al artículo original, empleando la misma arquitectura pero aplicada a tareas más específicas.

El algoritmo \ref{alg:codificadores-decodificadores} detalla la implementación de lo que sería un transformador de secuencia-a-secuencia.

\begin{algorithm}[H]
	\caption{$\boldsymbol{P} \leftarrow \text{transformadorCD}(\boldsymbol{x} \mid \boldsymbol{\theta})$}
	\begin{algorithmic}[1]
		\State \textbf{Entrada:} $\boldsymbol{z}, \boldsymbol{x} \in V^{*}$, dos secuencias de identificadores de tokens
		\State \textbf{Salida:} $\boldsymbol{P} \in (0,1)^{N_V \times \text{longitud}(x)}$, t de \textbf{P} representa $\hat{P_{\theta}}(x[t+1] \mid \boldsymbol{x}[1:t], \boldsymbol{z})$
		\State \textbf{Hiperparámetros:} $\ell_{max}, L_{enc}, L_{dec}, H, d_e, d_{mlp}, d_f \in \mathbb{N}$
		\State \textbf{Parámetros:} $\boldsymbol{\theta}$ que incluye a: 
		\State \hspace{0.5cm} $\boldsymbol{W_e} \in \mathbb{R}^{d_e \times N_V}, \boldsymbol{W_p} \in \mathbb{R}^{d_e \times \ell_{max}}$, matrices de embedding de token y posición
		\State \hspace{0.5cm} Para $l \in [L_{enc}]$:
		\State \hspace{1cm} $\boldsymbol{W}_{qkv,l}^{enc}$, auto-atención de la capa $l$
		\State \hspace{1cm} $\boldsymbol{\gamma}_l^1$, $\boldsymbol{\beta}_l^1$, $\boldsymbol{\gamma}_l^2$, $\boldsymbol{\beta}_l^2$ $\in \mathbb{R}^{d_e}$, dos conjuntos de normalización de capas
		\State \hspace{1cm} $\boldsymbol{W}_{mlp1}^l \in \mathbb{R}^{d_{mlp} \times d_e}$, $\boldsymbol{b}_{mlp1}^l \in \mathbb{R}^{d_{mlp}}$, $\boldsymbol{W}_{mlp2}^l \in \mathbb{R}^{d_{mlp} \times d_e}$, $\boldsymbol{b}_{mlp2}^l \in \mathbb{R}^{d_{mlp}}$
		\State \hspace{0.5cm} Para $l \in [L_{dec}]$:
		\State \hspace{1cm} $\boldsymbol{W}_{qkv,l}^{dec}$, $\boldsymbol{W}_{qkv,l}^{e/d}$, auto-atención y auto-atención cruzada de la capa $l$
		\State \hspace{1cm} $\boldsymbol{\gamma}_l^3$, $\boldsymbol{\beta}_l^3$, $\boldsymbol{\gamma}_l^4$, $\boldsymbol{\beta}_l^4$, $\boldsymbol{\gamma}_l^5$, $\boldsymbol{\beta}_l^5$  $\in \mathbb{R}^{d_e}$, tres conjuntos de normalización de capas
		\State \hspace{1cm} $\boldsymbol{W}_{mlp3}^l \in \mathbb{R}^{d_{mlp} \times d_e}$, $\boldsymbol{b}_{mlp3}^l \in \mathbb{R}^{d_{mlp}}$, $\boldsymbol{W}_{mlp4}^l \in \mathbb{R}^{d_{mlp} \times d_e}$, $\boldsymbol{b}_{mlp4}^l \in \mathbb{R}^{d_{mlp}}$
		\State \hspace{0.5cm} $\boldsymbol{W}_u \in \mathbb{R}^{N_V \times d_e}$, matriz inversa de embedding
		\Statex
		\Statex /* Codificación de la secuencia de contexto */
		\State $\ell_z \leftarrow \text{longitud}(z)$
		\State Para $t \in [\ell_z] : \boldsymbol{e}_t \leftarrow \boldsymbol{W}_e[:,z[t]] + \boldsymbol{W}_p[:,t]$
		\State $\boldsymbol{Z} \leftarrow [\boldsymbol{e}_1, \boldsymbol{e}_2, ..., \boldsymbol{e}_{\ell_z}]$
		\State \textbf{para} $\ell = 1, 2, ..., L_{enc}$ \textbf{hacer}
		\State \hspace{0.5cm} $\boldsymbol{Z} \leftarrow \boldsymbol{Z} + \text{atencion\_multicabezal}(\boldsymbol{Z} \mid \boldsymbol{W}_{qkv, l}^{enc}, \text{Mask} \equiv 1)$
		\State \hspace{0.5cm} Para $t \in [\ell_Z] : \boldsymbol{Z}[:,t] \leftarrow \text{normalizacion}(\boldsymbol{Z}[:,t] \mid \boldsymbol{\gamma}_l^1, \boldsymbol{\beta}_l^1)$
		\State \hspace{0.5cm} $\boldsymbol{Z} \leftarrow \boldsymbol{Z} + \boldsymbol{W}_{mlp2}^l \text{ReLU}(\boldsymbol{W}_{mlp1}^l \boldsymbol{Z} + \boldsymbol{b}_{mlp1}^l \boldsymbol{1}^T) + \boldsymbol{b}_{mlp2}^l \boldsymbol{1}^T$
		\State \hspace{0.5cm} Para $t \in [\ell_Z] : \boldsymbol{Z}[:,t] \leftarrow \text{normalizacion}(\boldsymbol{Z}[:,t] \mid \boldsymbol{\gamma}_l^2, \boldsymbol{\beta}_l^2)$
		\State \textbf{fin para}
		\Statex
		\Statex /* Decodificación de la primera secuencia condicionada por el contexto */
		\State $\ell_X \leftarrow \text{longitud}(x)$
		\State Para $t \in [\ell_X] : \boldsymbol{e}_t \leftarrow \boldsymbol{W}_e[:,x[t]] + \boldsymbol{W}_p[:,t]$
		\State $\boldsymbol{X} \leftarrow [\boldsymbol{e}_1, \boldsymbol{e}_2, ..., \boldsymbol{e}_{\ell_X}]$
		\State \textbf{para} $i = 1, 2, ..., L_{dec}$ \textbf{hacer}
		\State \hspace{0.5cm} $\boldsymbol{X} \leftarrow \boldsymbol{X} + \text{atencion\_multicabezal}(\boldsymbol{X} \mid \boldsymbol{W}_{qkv, l}^{dec}, \text{Mask}[t,t'] = [t \le t'])$
		\State \hspace{0.5cm} Para $t \in [\ell_X] : \boldsymbol{X}[:,t] \leftarrow \text{normalizacion}(\boldsymbol{X}[:,t] \mid \boldsymbol{\gamma}_l^3, \boldsymbol{\beta}_l^3)$
		\State \hspace{0.5cm} $\boldsymbol{X} \leftarrow \boldsymbol{X} + \text{atencion\_multicabezal}(\boldsymbol{X}, \boldsymbol{Z} \mid \boldsymbol{W}_{qkv, l}^{e/d}, \text{Mask} \equiv 1)$
		\State \hspace{0.5cm} Para $t \in [\ell_X] : \boldsymbol{X}[:,t] \leftarrow \text{normalizacion}(\boldsymbol{X}[:,t] \mid \boldsymbol{\gamma}_l^4, \boldsymbol{\beta}_l^4)$
		\State \hspace{0.5cm} $\boldsymbol{X} \leftarrow \boldsymbol{X} + \boldsymbol{W}_{mlp4}^l \text{ReLU}(\boldsymbol{W}_{mlp3}^l \boldsymbol{X} + \boldsymbol{b}_{mlp3}^l \boldsymbol{1}^T) + \boldsymbol{b}_{mlp4}^l \boldsymbol{1}^T$
		\State \textbf{fin para}
		\State \textbf{return} $P = \text{softmax}(\boldsymbol{W}_u \boldsymbol{X})$
	\end{algorithmic}
\label{alg:codificadores-decodificadores}	
\end{algorithm}

\begin{figure}[ht!]
	\centering
	\includegraphics{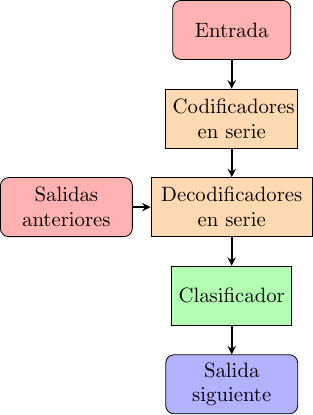}
	\label{fig:sec-a-sec}
	\caption{Arquitectura típica del modelo de transformador de secuencia-a-secuencia}
\end{figure}

En la Figura~\ref{fig:sec-a-sec} se muestra el diagrama funcional representativo de esta arquitectura. Este diagrama es equivalente al de la Figura~\ref{fig:transformador}, donde se visualizan los elementos constitutivos a un nivel alto de abstracción. Los componentes clave que se destacan incluyen: los bloques encargados del tratamiento de la secuencia de entrada, los codificadores, los decodificadores y, finalmente, la capa de clasificación y salida.

\begin{figure}[ht!]
	\centering
	\includegraphics{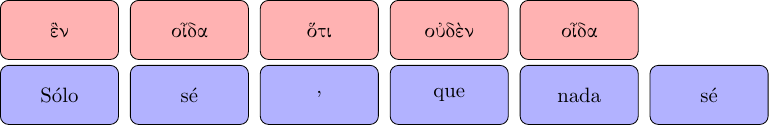}
	\label{fig:sec-a-sec-ejemplo}
	\caption{Ejemplo de un conjunto de entrada y salida común de un modelo secuencia-a-secuencia. En este caso sería un modelo de traducción de griego clásico al español}
\end{figure}

En la Figura~\ref{fig:sec-a-sec-ejemplo} se presenta un ejemplo de una secuencia típica de entrada y su correspondiente secuencia de salida. Las secuencias mostradas ilustran un caso de traducción, donde la secuencia de entrada en griego clásico se traduce a una secuencia en español. El modelo recibe como entrada la primera secuencia y produce como salida la segunda.

En el algoritmo \ref{alg:entrenamiento-codificadores-decodificadores} se presenta el código necesario para llevar a cabo la optimización del modelo utilizando un conjunto de datos de entrada.

\begin{algorithm}[H]
	\caption{$\hat{\boldsymbol{\theta}} \leftarrow \texttt{entrenamientoCD}(\boldsymbol{z}_{1:N_{data}}, \boldsymbol{x}_{1:N_{data}}, \boldsymbol{\theta})$}
	\begin{algorithmic}[1]
		\State \textbf{Entrada:} $(\boldsymbol{z}_n, \boldsymbol{x}_n)_{n=1}^{N_{data}}$, un conjunto de datos de parejas de secuencias
		\State \textbf{Entrada:} $\boldsymbol{\theta}$, valores iniciales de los parámetros del transformador
		\State \textbf{Salida:} $\hat{\boldsymbol{\theta}}$, valores de los parámetros después del entrenamiento
		\State \textbf{Hiperparámetros:} $N_{epochs} \in \mathbb{N}$, tasa de aprendizaje $\eta \in (0, \infty)$
		\State
		\For{$i = 1, 2, \dots, N_{epochs}$}
		\For{$n = 1, 2, \dots, N_{data}$}
		\State $\ell \leftarrow \text{longitud}(\boldsymbol{x}_n)$
		\State $\boldsymbol{P_{\theta}} \leftarrow \texttt{transformador\_de\_codificadores\_decodificadores}(\boldsymbol{z}_n, \boldsymbol{x}_n \mid \boldsymbol{\theta})$
		\State $loss(\boldsymbol{\theta}) \leftarrow - \sum_{t=1}^{\ell-1} \log P(\boldsymbol{\theta})[x_n[t+1], t]$
		\State $\boldsymbol{\theta} \leftarrow \boldsymbol{\theta} - \eta \nabla \, loss(\boldsymbol{\theta})$
		\EndFor
		\EndFor
		\State \textbf{Retorna:} $\hat{\boldsymbol{\theta}} = \boldsymbol{\theta}$
	\end{algorithmic}
	\label{alg:entrenamiento-codificadores-decodificadores}
\end{algorithm}

\subsection{Transformadores bidireccionales o de auto-codificación}

Los modelos de auto-codificación son aquellos que están pre-entrenados para corregir entradas corrompidas artificialmente. Este proceso tiene como objetivo entrenar al modelo en la reconstrucción del texto original. Este tipo de modelos corresponde a la parte del codificador de la arquitectura original, ya que no utilizan máscaras para anular el efecto del texto posterior. Los modelos de auto-codificación construyen una representación bidireccional de la frase en su totalidad y pueden ser posteriormente entrenados para resolver tareas específicas, como la generación de texto, aunque su aplicación principal es la clasificación de frases. Un ejemplo típico de estos modelos es \textbf{BERT} \cite{devlin2018bert}.

El algoritmo \ref{alg:codificadores} recoge los detalles de implementación de lo que sería un transformador bidireccional, es decir, un transformador compuesto únicamente por codificadores.

\begin{algorithm}[H]
	\caption{$\boldsymbol{P} \leftarrow codificador (\boldsymbol{x} \mid \boldsymbol{\theta})$}
	\begin{algorithmic}
		\State /* Transformador formado por codificadores, bidireccional o auto-codificador */
		\State \textbf{Entrada:} $\boldsymbol{x} \in V^{*}$, una secuencia de identificadores de tokens
		\State \textbf{Salida:} $\boldsymbol{P} \in (0,1)^{N_V \times \text{longitud}(x)}$, donde cada columna de $\boldsymbol{P}$ es una distribución sobre todo el vocabulario
		\State \textbf{Hiperparámetros:} $\ell_{max}, L, H, d_e, d_{mlp}, d_f \in \mathbb{N}$
		\State \textbf{Parámetros:} $\boldsymbol{\theta}$ que incluye:
		\State \hspace{0.5cm} $\boldsymbol{W_e} \in \mathbb{R}^{d_e \times N_V}$, $\boldsymbol{W_p} \in \mathbb{R}^{d_e \times \ell_{max}}$, matrices de embedding de token y posición
		\State \hspace{0.5cm} Para $l \in [L]$:
		\State \hspace{1cm} $\boldsymbol{W}_{qkv,l}$, parámetros de auto-atención de la capa $l$
		\State \hspace{1cm} $\boldsymbol{\gamma}_l^1$, $\boldsymbol{\beta}_l^1$, $\boldsymbol{\gamma}_l^2$, $\boldsymbol{\beta}_l^2 \in \mathbb{R}^{d_e}$, normalización de capas
		\State \hspace{1cm} $\boldsymbol{W}_{mlp1}^l \in \mathbb{R}^{d_{mlp} \times d_e}$, $\boldsymbol{b}_{mlp1}^l \in \mathbb{R}^{d_{mlp}}$, $\boldsymbol{W}_{mlp2}^l \in \mathbb{R}^{d_{mlp} \times d_e}$, $\boldsymbol{b}_{mlp2}^l \in \mathbb{R}^{d_{mlp}}$
		\State \hspace{0.5cm} $\boldsymbol{W}_f \in \mathbb{R}^{d_f \times d_e}$, $\boldsymbol{b}_f \in \mathbb{R}^{d_f}$, $\boldsymbol{\gamma}, \boldsymbol{\beta} \in \mathbb{R}^{d_f}$, proyección lineal y normalización final
		\State \hspace{0.5cm} $\boldsymbol{W}_u \in \mathbb{R}^{N_V \times d_e}$, matriz inversa de embedding
		\State
		\begin{tabbing}
			\textbf{1} \hspace{0.5cm} \= $\ell \leftarrow \text{longitud}(x)$ \\
			\textbf{2} \hspace{0.5cm} \= \textbf{para} $t \in [\ell]$ \textbf{hacer} \\
			\hspace{1cm} \= $\boldsymbol{e}_t \leftarrow \boldsymbol{W}_e[:,x[t]] + \boldsymbol{W}_p[:,t]$ \\
			\textbf{3} \hspace{0.5cm} \= $\boldsymbol{X} \leftarrow [\boldsymbol{e}_1, \boldsymbol{e}_2, ..., \boldsymbol{e}_\ell]$ \\
			\textbf{4} \hspace{0.5cm} \= \textbf{para} $l = 1, 2, ..., L$ \textbf{hacer} \\
			\hspace{1cm} \= $\boldsymbol{X} \leftarrow \boldsymbol{X} + \text{atencion\_multicabezal}(\boldsymbol{X} \mid \boldsymbol{W}_{qkv,l}, \text{Mask} \equiv 1)$ \\
			\textbf{5} \hspace{0.5cm} \= \textbf{para} $t \in [\ell]$ \textbf{hacer} \\
			\hspace{1cm} \= $\boldsymbol{X}[:,t] \leftarrow \text{normalizacion}(\boldsymbol{X}[:,t] \mid \boldsymbol{\gamma}_l^1, \boldsymbol{\beta}_l^1)$ \\
			\textbf{6} \hspace{0.5cm} \= $\boldsymbol{X} \leftarrow \boldsymbol{X} + \boldsymbol{W}_{mlp2}^l \, \text{GELU}(\boldsymbol{W}_{mlp1}^l \boldsymbol{X} + \boldsymbol{b}_{mlp1}^l \boldsymbol{1}^T) + \boldsymbol{b}_{mlp2}^l \boldsymbol{1}^T$ \\
			\textbf{7} \hspace{0.5cm} \= \textbf{para} $t \in [\ell]$ \textbf{hacer} \\
			\hspace{1cm} \= $\boldsymbol{X}[:,t] \leftarrow \text{normalizacion}(\boldsymbol{X}[:,t] \mid \boldsymbol{\gamma}_l^2, \boldsymbol{\beta}_l^2)$ \\
			\textbf{8} \hspace{0.5cm} \= \textbf{fpara} \\
			\textbf{9} \hspace{0.5cm} \= $\boldsymbol{X} \leftarrow \text{GELU}(\boldsymbol{W}_f \boldsymbol{X} + \boldsymbol{b}_f \boldsymbol{1}^T)$ \\
			\textbf{10} \hspace{0.5cm} \= \textbf{para} $t \in [\ell]$ \textbf{hacer} \\
			\hspace{1cm} \= $\boldsymbol{X}[:,t] \leftarrow \text{normalizacion}(\boldsymbol{X}[:,t] \mid \boldsymbol{\gamma}, \boldsymbol{\beta})$ \\
			\textbf{11} \hspace{0.5cm} \= \textbf{fpara} \\
			\textbf{12} \hspace{0.5cm} \= \textbf{return} $P = \text{softmax}(\boldsymbol{W}_u \boldsymbol{X})$
		\end{tabbing}
	\end{algorithmic}
	\label{alg:codificadores}
\end{algorithm}

\begin{algorithm}[H]
	\caption{$\hat{\boldsymbol{\theta}} \leftarrow \text{entrenamientoC}(\boldsymbol{x}_{1:N_{data}}, \boldsymbol{\theta})$}
	\begin{algorithmic}[1]
		\State \textbf{Entrada:} $\{\boldsymbol{x}_n\}_{n=1}^{N_{data}}$, un conjunto de datos de secuencias
		\State \textbf{Entrada:} $\boldsymbol{\theta}$, valores de partida de los parámetros del transformador
		\State \textbf{Salida:} $\hat{\boldsymbol{\theta}}$, valor de los parámetros después del entrenamiento
		\State \textbf{Hiperparámetros:} $N_{epochs} \in \mathbb{N}, \eta \in (0, \infty), p_{mask} \in (0,1)$
		\State
		\For{$i = 1, 2, \dots, N_{epochs}$}
		\For{$n = 1, 2, \dots, N_{data}$}
		\State $\ell \leftarrow \text{longitud}(\boldsymbol{x}_n)$
		\For{$t = 1, 2, \dots, \ell$}
		\State $\tilde{x}_n[t] \leftarrow \text{enmascara\_token}$ o $x_n[t]$ aleatoriamente con probabilidad $p_{mask}$ o $1 - p_{mask}$
		\EndFor
		\State $\tilde{T} \leftarrow \{t \in [\ell] : \tilde{x}_n[t] = \text{enmascara\_token}\}$
		\State $\boldsymbol{P}(\boldsymbol{\theta}) \leftarrow \text{transformador\_de\_codificadores}(\tilde{\boldsymbol{x}}_n \mid \boldsymbol{\theta})$
		\State $loss(\boldsymbol{\theta}) = - \sum_{t \in \tilde{T}} \log P(\boldsymbol{\theta})[x_n[t],t]$
		\State $\boldsymbol{\theta} \leftarrow \boldsymbol{\theta} - \eta \nabla loss(\boldsymbol{\theta})$
		\EndFor
		\EndFor
		\State \textbf{Retorna:} $\hat{\boldsymbol{\theta}} = \boldsymbol{\theta}$
	\end{algorithmic}
	\label{alg:entrenamiento-codificadores}
\end{algorithm}

La red de codificación apila en serie $L$ bloques del mismo tipo. La salida de cada bloque sirve como entrada para el siguiente. Las dimensiones de la transformación se mantienen constantes, de modo que la entrada y la salida de cada bloque tienen el mismo tamaño.

\begin{figure}[ht!]
	\centering
	\includegraphics{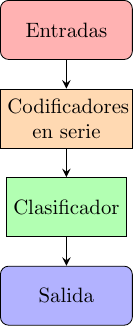}
	\label{fig:autocodificadores}
	\caption{Arquitectura típica del modelo de transformador auto-codificador}
\end{figure}

\begin{figure}[ht!]
	\centering
	\includegraphics{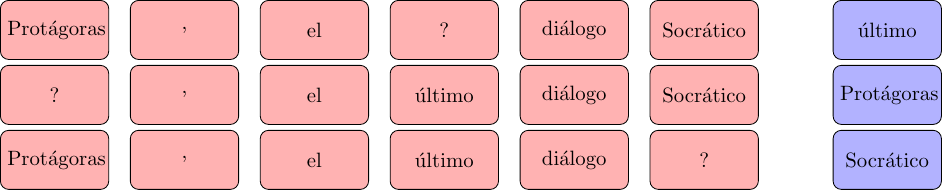}
	\label{fig:autocodificacion-ejemplo}
	\caption{Ejemplo de entradas y salidas típicas de los modelos de auto-codificación. En este caso se presentan tres entradas posibles con sus valores de salida esperados}
\end{figure}

\begin{table}[ht!]
	\centering
	\begin{tabular}{ l | c | c | c | c | c | c }
		\hline
		\textbf{} & \textbf{BERT-b} & \textbf{BERT-l} & \textbf{XLNet-l} & \textbf{RoBERTa-l} & \textbf{DistillBERT-b} & \textbf{ALBERT-l} \\
		\hline
		\multicolumn{7}{c}{\textbf{Arquitectura}} \\
		\hline
		\textbf{$n_{param}$} & \SI{110e6}{} & \SI{340e6}{} & \SI{340e6}{} & \SI{355e6}{} & \SI{66e6}{} & \SI{18e6}{} \\
		\hline
		\textbf{$T_{param}$} & \SI{432}{MB} & \SI{1360}{MB} & \SI{1360}{MB} & \SI{1420}{MB} & \SI{264}{MB} & \SI{72}{MB} \\
		\hline
		\textbf{$\ell_{max}$} & 512 & 512 & 512 & 512 & 512 & 512 \\
		\hline
		\textbf{$L_{enc}$} & 12 & 24 & 24 & 24 & 6 & 12 \\
		\hline
		\textbf{$H$} & 12 & 16 & 16 & 16 & 12 & 12 \\
		\hline
		\textbf{$d_{attn}$} & 64 & 64 & 64 & 64 & 64 & 128 \\
		\hline
		\textbf{$d_{e}$} & 768 & 1024 & 1024 & 1024 & 768 & 128 \\
		\hline
		\textbf{$d_{mlp}$} & 3072 & 4096 & 4096 & 4096 & 3072 & 512 \\
		\hline
		\textbf{$N_V$} & 30522 & 30522 & 30522 & 30522 & 30522 & 30000 \\
		\hline
		\textbf{$c_{entrada}$} & \multicolumn{2}{c}{WordPiece} & WordPiece & WordPiece & WordPiece & SentencePiece \\
		\hline
		\multicolumn{7}{c}{\textbf{Entrenamiento}} \\
		\hline
		\textbf{$T_{datos}$} & \SI{16}{GB} & \SI{16}{GB} & \SI{113}{GB} & \SI{160}{GB} & \SI{16}{GB} & \SI{16}{GB} \\
		\hline
		\textbf{$n_{tokens}$} & \SI{3.3e9}{} & \SI{3.3e9}{} & \SI{33e9}{} & \SI{33e9}{} & \SI{3.3e9}{} & \SI{3.3e9}{} \\
		\hline
		\textbf{$ratio_{compr}$} & 30.5 & 9.7 & 97.0 & 100.0 & 50.0 & 20.0 \\
		\hline
		\textbf{Optimizador} & Adam & Adam & Adam & Adam & Adam & Adam \\
		\hline
		\textbf{BS} & 256 & 256 & 512 & 8000 & 512 & 256 \\
		\hline
		\multicolumn{7}{c}{\textbf{Rendimiento}} \\
		\hline
		\textbf{GLUE} & base & +15\% & +20\% & -3\% & +25\% & +10\% \\
		\hline
	\end{tabular}
	\caption{Características de algunos de los primeros modelos bidireccionales más conocidos.}
	\label{table:bidireccionales}
\end{table}

\subsection{Transformadores auto-regresivos}

Los modelos \textit{auto-regresivos} son aquellos que, a partir de una secuencia de entrada inicial, generan nuevas salidas que, a su vez, se incorporan a la secuencia de entrada para generar subsecuentes salidas. Su principal aplicación es la \textit{generación de texto}, aunque también pueden ajustarse mediante entrenamiento posterior para adaptarlos a la resolución de problemas específicos. Un ejemplo destacado de este tipo de modelos es la familia de \textbf{GPT}.

El algoritmo \ref{alg:decodificadores} detalla la implementación de un transformador auto-regresivo, el cual está compuesto únicamente por decodificadores.

\begin{algorithm}[h]
	\caption{$\boldsymbol{P} \leftarrow \text{transformadorD}(\boldsymbol{x} \mid \boldsymbol{\theta})$}
	\begin{algorithmic}
		\State \textbf{/* Transformador formado solo por decodificadores (auto-regresivo) */}
		\State \textbf{Entrada:} $\boldsymbol{x} \in V^{*}$, secuencia de identificadores de tokens
		\State \textbf{Salida:} $\boldsymbol{P} \in (0,1)^{N_V \times \text{longitud}(x)}$, donde la columna $t$ de $\boldsymbol{P}$ representa $\hat{P_{\theta}}(x[t+1] \mid \boldsymbol{x}[1:t])$
		\State \textbf{Hiperparámetros:} $\ell_{\text{max}}, L, H, d_e, d_{mlp}, d_f \in \mathbb{N}$
		\State \textbf{Parámetros:} $\boldsymbol{\theta}$ que incluye:
		\State \hspace{0.5cm} $\boldsymbol{W_e} \in \mathbb{R}^{d_e \times N_V}, \boldsymbol{W_p} \in \mathbb{R}^{d_e \times \ell_{\text{max}}}$, matrices de embedding de token y posición
		\State \hspace{0.5cm} Para $l \in [L]$:
		\State \hspace{1cm} $\boldsymbol{W}_{qkv,l}$, parámetros de auto-atención para la capa $l$
		\State \hspace{1cm} $\boldsymbol{\gamma}_l^1$, $\boldsymbol{\beta}_l^1$, $\boldsymbol{\gamma}_l^2$, $\boldsymbol{\beta}_l^2$ $\in \mathbb{R}^{d_e}$, conjuntos de normalización de capas
		\State \hspace{1cm} $\boldsymbol{W}_{mlp1}^l \in \mathbb{R}^{d_{mlp} \times d_e}$, $\boldsymbol{b}_{mlp1}^l \in \mathbb{R}^{d_{mlp}}$, $\boldsymbol{W}_{mlp2}^l \in \mathbb{R}^{d_{mlp} \times d_e}$, $\boldsymbol{b}_{mlp2}^l \in \mathbb{R}^{d_{mlp}}$
		\State \hspace{0.5cm} $\boldsymbol{W}_f \in \mathbb{R}^{d_f \times d_e}$, $\boldsymbol{b}_f \in \mathbb{R}^{d_f}$, $\boldsymbol{\gamma}, \boldsymbol{\beta} \in \mathbb{R}^{d_f}$, proyección lineal y normalización final
		\State \hspace{0.5cm} $\boldsymbol{W}_u \in \mathbb{R}^{N_V \times d_e}$, matriz de embedding inverso
		\State
		\begin{tabular}{ l l }
			\textbf{1} & $\ell \leftarrow \text{longitud}(x)$ \\
			\textbf{2} & Para $t \in [\ell] : \boldsymbol{e}_t \leftarrow \boldsymbol{W}_e[:, x[t]] + \boldsymbol{W}_p[:, t]$ \\
			\textbf{3} & $\boldsymbol{X} \leftarrow [\boldsymbol{e}_1, \boldsymbol{e}_2, \dots, \boldsymbol{e}_\ell]$ \\
			\textbf{4} & \textbf{Para} $\ell = 1, 2, \dots, L$ \textbf{hacer} \\
			\textbf{5} & \hspace{0.5cm} Para $t \in [\ell] : \boldsymbol{X}[:, t] \leftarrow \text{normalizacion}(\boldsymbol{X}[:, t] \mid \boldsymbol{\gamma}_l^1, \boldsymbol{\beta}_l^1)$ \\
			\textbf{6} & \hspace{0.5cm} $\boldsymbol{X} \leftarrow \boldsymbol{X} + \text{atencion\_multicabezal}(\boldsymbol{X} \mid \boldsymbol{W}_{qkv, l}, \text{Mask}[t,t'] = [[t \leq t']])$ \\
			\textbf{7} & \hspace{0.5cm} Para $t \in [\ell] : \boldsymbol{X}[:, t] \leftarrow \text{normalizacion}(\boldsymbol{X}[:, t] \mid \boldsymbol{\gamma}_l^2, \boldsymbol{\beta}_l^2)$ \\
			\textbf{8} & \hspace{0.5cm} $\boldsymbol{X} \leftarrow \boldsymbol{X} + \boldsymbol{W}_{mlp2}^l \cdot \text{GELU}(\boldsymbol{W}_{mlp1}^l \boldsymbol{X} + \boldsymbol{b}_{mlp1}^l \boldsymbol{1}^T) + \boldsymbol{b}_{mlp2}^l \boldsymbol{1}^T$ \\
			\textbf{9} & \hspace{0.5cm} \textbf{fpara} \\
			\textbf{10} & Para $t \in [\ell] : \boldsymbol{X}[:, t] \leftarrow \text{normalizacion}(\boldsymbol{X}[:, t] \mid \boldsymbol{\gamma}, \boldsymbol{\beta})$ \\
			\textbf{11} & \textbf{return} $P = \text{softmax}(\boldsymbol{W}_u \cdot \boldsymbol{X})$ \\
		\end{tabular}
	\end{algorithmic}
	\label{alg:decodificadores}
\end{algorithm}

Igual que en los codificadores, la red de decodificación apila en serie \( N \) elementos del mismo tipo. La salida de cada bloque sirve como entrada para el siguiente. Las dimensiones de la transformación se mantienen constantes, por lo que la entrada y la salida tienen el mismo tamaño.

\begin{figure}[ht]
	\centering
	\includegraphics{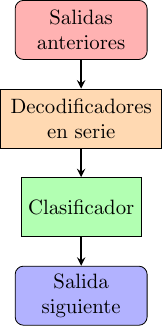}
	\label{fig:autoregresivo}
	\caption{Arquitectura de transformador auto-regresivo}
\end{figure}

\begin{figure}[ht]
	\centering
	\includegraphics{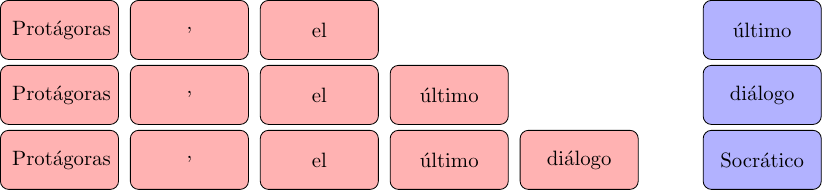}
	\label{fig:autoregresivo-ejemplo}
	\caption{Ejemplo de entradas y salidas típicas de los modelos auto-regresivos. En este caso se presentan tres entradas posibles con sus respectivas salidas esperadas}
\end{figure}

\begin{table}[ht]
	\centering
	\begin{tabular}{@{}lccc@{}}
		\toprule
		& \textbf{GPT-1} & \textbf{GPT-2} & \textbf{GPT-3} \\ \midrule
		\multicolumn{4}{c}{\textbf{Arquitectura}} \\ \midrule
		\textbf{$n_{\text{param}}$} & \num{1.17e8} & \num{1.5e9} & \num{1.75e11} \\
		\textbf{$T_{\text{param}}$} & \SI{468}{\mega\byte} & \SI{6}{\giga\byte} & \SI{700}{\giga\byte} \\
		\textbf{$\ell_{\text{max}}$} & 512 & 1024 & 2048 \\
		\textbf{$L_{\text{dec}}$} & 12 & 48 & 96 \\
		\textbf{$H$} & 12 & 12 & 96 \\
		\textbf{$d_{\text{attn}}$} & 64 & - & 128 \\
		\textbf{$d_{\text{mlp}}$} & 3072 & - & - \\
		\textbf{$N_V$} & 40478 & 50257 & 50257 \\
		\textbf{$c_{\text{entrada}}$} & BPE & BPE & BPE \\
		\textbf{$d_e$} & 768 & 1600 & 12288 \\ \midrule
		\multicolumn{4}{c}{\textbf{Entrenamiento}} \\ \midrule
		\textbf{$T_{\text{datos}}$} & \SI{1}{\giga\byte} & \SI{40}{\giga\byte} & \SI{2}{\tera\byte} \\
		\textbf{$n_{\text{tokens}}$} & \num{2.5e8} & \num{1e10} & \num{4.99e11} \\
		\textbf{$\text{ratio}_{\text{compr}}$} & 2.13 & 6.66 & 2.85 \\
		\textbf{Optimizador} & Adam & Adam & Adam \\
		\textbf{BS} & 64 & 512 & \num{3.2e6} \\ \bottomrule
	\end{tabular}
	\caption{Características de los modelos autoregresivos de la familia GPT.}
	\label{table:autoregresivos}
\end{table}

En el algoritmo \ref{alg:entrenamiento-decodificadores} se muestra el conjunto de operaciones que se llevan a cabo en un bucle de entrenamiento.

\begin{algorithm}[H]
	\caption{$\hat{\boldsymbol{\theta}} \leftarrow \text{entrenamientoD}(\boldsymbol{x}_{1:N_{\text{data}}}, \boldsymbol{\theta})$}
	\begin{algorithmic}[1]
		\State \textbf{Entrada:} $\{\boldsymbol{x}_n\}_{n=1}^{N_{\text{data}}}$, un conjunto de datos de secuencias
		\State \textbf{Entrada:} $\boldsymbol{\theta}$, valores iniciales de los parámetros del transformador
		\State \textbf{Salida:} $\hat{\boldsymbol{\theta}}$, parámetros después del entrenamiento
		\State
		\For{$i = 1, 2, ..., N_{\text{epochs}}$}
		\For{$n = 1, 2, ..., N_{\text{data}}$}
		\State $\ell \leftarrow \text{longitud}(\boldsymbol{x}_n)$
		\State $\boldsymbol{P_{\theta}} \leftarrow \text{transformador\_de\_decodificadores}(\boldsymbol{x}_n \mid \boldsymbol{\theta})$
		\State $loss(\boldsymbol{\theta}) = - \sum_{t=1}^{\ell-1} \log P (\boldsymbol{\theta})[x_n[t+1], t]$
		\State $\boldsymbol{\theta} \leftarrow \boldsymbol{\theta} - \eta \nabla_{\boldsymbol{\theta}} \, loss(\boldsymbol{\theta})$
		\EndFor
		\EndFor
		\State \textbf{Retorna:} $\hat{\boldsymbol{\theta}} = \boldsymbol{\theta}$
	\end{algorithmic}
	\label{alg:entrenamiento-decodificadores}
\end{algorithm}

\begin{algorithm}[H]
	\caption{$\hat{\boldsymbol{\theta}} \leftarrow \text{inferencia\_transformadorD}(\boldsymbol{x}, \hat{\boldsymbol{\theta}})$}
	\begin{algorithmic}[1]
		\State /* Predicción usando un modelo entrenado */
		\State \textbf{Entrada:} $\hat{\boldsymbol{\theta}}$, parámetros entrenados
		\State \textbf{Entrada:} $\boldsymbol{x} \in V^{*}$, una secuencia de entrada
		\State \textbf{Salida:} $y \in V^{*}$, predicción de la continuación de la secuencia de entrada
		\State \textbf{Hiperparámetros:} $\ell_{\text{gen}} \in \mathbb{N}$, $\tau \in (0, \infty)$
		\State
		\For{$n = 1, 2, ..., \ell_{\text{gen}}$}
		\State $\boldsymbol{P} \leftarrow \text{transformador\_de\_decodificadores}(\boldsymbol{x} \mid \hat{\boldsymbol{\theta}})$
		\State $\boldsymbol{p} \leftarrow \boldsymbol{P}[:, \ell + n - 1]$
		\State Muestrea un token $y$ de $\boldsymbol{q} \sim \boldsymbol{p}^{1/\tau}$
		\State $\boldsymbol{x} \leftarrow [\boldsymbol{x}, y]$
		\EndFor
		\State \textbf{Retorna:} $y = \boldsymbol{x}[\ell + 1 : \ell + \ell_{\text{gen}}]$
	\end{algorithmic}
	\label{alg:inferencia-decodificadores}
\end{algorithm}

\begin{algorithm}[H]
	\caption{$\hat{\boldsymbol{\theta}} \leftarrow \text{inferencia\_transformadorCD}(\boldsymbol{z}, \hat{\boldsymbol{\theta}})$}
	\begin{algorithmic}[1]
		\State /* Predicción usando un transformador secuencia-a-secuencia */
		\State \textbf{Entrada:} $\hat{\boldsymbol{\theta}}$, parámetros entrenados
		\State \textbf{Entrada:} $\boldsymbol{z} \in V^{*}$, una secuencia de entrada
		\State \textbf{Salida:} $\hat{\boldsymbol{x}} \in V^{*}$, predicción de la secuencia de salida
		\State \textbf{Hiperparámetros:} $\tau \in (0, \infty)$
		\State
		\State $\hat{\boldsymbol{x}} \leftarrow [\texttt{bos\_token}]$ \text{ // Inicializa la secuencia de salida con el token de inicio}
		\State $y \leftarrow 0$ \text{ // Inicializa el token de salida}
		\While{$y \neq \texttt{eos\_token}$} \text{ // Mientras no se alcance el token de fin de secuencia}
		\State $\boldsymbol{P} \leftarrow \text{transformador\_CD}(\boldsymbol{z}, \hat{\boldsymbol{x}} \mid \hat{\boldsymbol{\theta}})$
		\State $\boldsymbol{p} \leftarrow \boldsymbol{P}[:, \text{longitud}(\hat{\boldsymbol{x}})]$
		\State Muestrea un token $y$ de $\boldsymbol{q} \sim \boldsymbol{p}^{1/\tau}$ \text{ // Muestreo usando la distribución suavizada}
		\State $\hat{\boldsymbol{x}} \leftarrow [\hat{\boldsymbol{x}}, y]$ \text{ // Añade el token a la secuencia de salida}
		\EndWhile
		\State \textbf{Retorna:} $\hat{\boldsymbol{x}}$ \text{ // Secuencia de salida predicha}
	\end{algorithmic}
	\label{alg:inferencia-decodificadores2}
\end{algorithm}

\section{Transformadores para procesamiento del lenguaje natural (NLP)}

\subsection{Desarrollos previos: embeddings}

Antes de la introducción de los algoritmos basados en transformadores, es relevante mencionar un conjunto de algoritmos desarrollados en años anteriores, concretamente aquellos destinados a la creación de \textit{embeddings}.

En el contexto de procesamiento de lenguaje natural (NLP, por sus siglas en inglés), un \textit{embedding} es una representación matemática del significado de las palabras. Esta transformación parte de la premisa de que es posible representar el conocimiento en un espacio vectorial multidimensional. De este modo, cada palabra del lenguaje se asigna a un punto en dicho espacio. Los vectores que representan a las palabras son conocidos como \textit{embeddings}.

Durante los años previos a la aparición de los modelos basados en transformadores, se llevaron a cabo diversas publicaciones que definieron distintos métodos para realizar esta transformación. Entre los más conocidos se encuentran \textbf{Word2Vec} \cite{word2vec}, \textbf{GloVe} \cite{pennington2014glove}, y \textbf{FastText} \cite{joulin2016fasttext}, entre otros.

Estos algoritmos realizan la transformación utilizando una red neuronal poco profunda. Entre sus propiedades más destacadas se encuentran las siguientes:

\begin{itemize}
	\item \textbf{Similaridad entre palabras sinónimas o semánticamente relacionadas:} En el espacio transformado, las palabras con significados similares tienden a ubicarse en posiciones cercanas. Por ejemplo, términos como \textit{coche}, \textit{vehículo} o \textit{furgoneta} se encuentran más próximos entre sí que respecto a palabras como \textit{luna}, \textit{espacio} o \textit{árbol}. Esta proximidad puede medirse mediante métricas como la distancia euclidiana o la similaridad del coseno.
	
	\item \textbf{Codificación de relaciones lingüísticas entre palabras:} Una propiedad notable de estas representaciones es que ciertas relaciones lingüísticas pueden expresarse como transformaciones lineales. Por ejemplo, la diferencia vectorial entre \textit{hombre} y \textit{mujer} es similar a la existente entre \textit{rey} y \textit{reina}, \textit{tío} y \textit{tía}, o \textit{actor} y \textit{actriz}, lo cual sugiere una transformación que captura el concepto de género. Esto permite realizar operaciones aproximadas como:
	\[
	\vec{\text{rey}} - \vec{\text{hombre}} + \vec{\text{mujer}} \approx \vec{\text{reina}}, \quad \vec{\text{París}} - \vec{\text{Francia}} + \vec{\text{Alemania}} \approx \vec{\text{Berlín}}
	\]
\end{itemize}

Las limitaciones de estos métodos derivan de la propia naturaleza de su diseño. Al estar entrenados para codificar el significado de las palabras de forma aislada, no incorporan información contextual. Una solución propuesta para integrar el contexto, dentro del entorno de las redes neuronales, consistió en utilizar los \textit{embeddings} como una etapa de preprocesamiento a la entrada de redes neuronales recurrentes (RNNs). El objetivo era que estas redes se encargaran de codificar el contexto para resolver tareas específicas de procesamiento del lenguaje natural.

Los transformadores (\textit{Transformers}) representan un avance significativo respecto a este enfoque. Su objetivo es permitir un aprendizaje de extremo a extremo, donde las representaciones vectoriales de las palabras (los \textit{embeddings}) pasan a formar parte de la propia arquitectura del modelo. Estas representaciones se aprenden conjuntamente con el resto de parámetros durante el entrenamiento, quedando integradas como parte de la red global. En particular, los \textit{embeddings} iniciales se combinan con los mecanismos de auto-atención desde la primera capa, lo que permite capturar relaciones contextuales desde el inicio del procesamiento.

\subsection{Inicios de los Transformadores para NLP}

Las publicaciones iniciales que marcaron la evolución de los transformadores se enmarcan en el ámbito del procesamiento del lenguaje natural (NLP). A continuación, se presenta un resumen cronológico de los trabajos que consideramos más relevantes para la definición de este paradigma de diseño, así como sus principales aportaciones:

\begin{itemize}
	\item \textbf{Introducción de la arquitectura} \cite{vaswani2017attention}: Este trabajo introduce la arquitectura \textit{Transformer}, basada exclusivamente en mecanismos de atención, eliminando completamente las estructuras recurrentes. Supuso un cambio de paradigma en el diseño de modelos para NLP, destacando por su capacidad de paralelización y eficiencia computacional.
	
	\item \textbf{ULMFiT (Universal Language Model Fine-tuning for Text Classification)} \cite{howard2018universal}: Los autores proponen un enfoque de aprendizaje por transferencia para resolver tareas de NLP, mediante el uso de modelos de lenguaje generales preentrenados con datos no supervisados y afinados posteriormente en tareas específicas. Introducen además técnicas que serían ampliamente adoptadas posteriormente, como los \textit{slanted triangular learning rates} (también conocidos como \textit{warm-up}).
	
	\item \textbf{GPT} \cite{radford2018improving} y \textbf{GPT-2} \cite{radford2019language} (Generative Pre-trained Transformers): Se propone una arquitectura compuesta únicamente por decodificadores, orientada a un entrenamiento autoregresivo. La principal aportación de estas publicaciones es demostrar la utilidad de entrenar modelos de gran escala sobre grandes cantidades de datos no supervisados, logrando resultados competitivos en tareas de NLP mediante un simple ajuste posterior.
	
	\item \textbf{BERT (Bidirectional Encoder Representations from Transformers)} \cite{devlin2018bert}: Este trabajo propone una arquitectura basada exclusivamente en codificadores, permitiendo aprender representaciones bidireccionales del texto. El modelo es preentrenado con tareas como \textit{masked language modeling} y \textit{next sentence prediction}, y posteriormente afinado con una capa de salida específica para cada tarea. Supuso un punto de inflexión en el rendimiento de los sistemas de NLP.
	
	\item \textbf{RoBERTa (Robustly Optimized BERT Approach)} \cite{liu2019roberta}: Los autores mantienen la arquitectura de BERT, pero introducen mejoras significativas en el procedimiento de entrenamiento, como el uso de un corpus más grande, la eliminación de la tarea de predicción de la siguiente frase y un mayor tiempo de entrenamiento. El resultado es un modelo más robusto y eficaz.
	
	\item \textbf{T5 (Text-to-Text Transfer Transformer)} \cite{raffel2020exploring}: Este trabajo reformula todas las tareas de NLP como problemas de transformación de texto a texto. Además, realiza un estudio exhaustivo del impacto de distintos factores (como la función objetivo, la arquitectura, los conjuntos de datos y las estrategias de transferencia) sobre el rendimiento del modelo en múltiples tareas de NLP. También propone un nuevo estándar para reportar resultados en términos textuales en lugar de métricas puramente numéricas.
\end{itemize}

%\subsection{Tipos de representaciones aprendidas por los Transformadores}

%Vamos a intentar clasificar los distintos tipos de representaciones que es posible conseguir dependiendo del tipo de transformador que se entrene.

%\begin{enumerate}
%	\item Representaciones contextuales vs no-contextuales: \textbf{Word2Vec} o \textbf{FastText} son no contextuales mientras que todos los procedentes delos transformadores son textuales.
%	\item Unidireccional vs Bidireccionales
%\end{enumerate}

%\subsection{Escalado de tamaño en las capacidades de los transformadores}

%Explicar lo de las emerging properties

\section{Transformadores para Visión (ViT)}

En las secciones anteriores hemos analizado el funcionamiento de los transformadores en su aplicación de referencia: el procesamiento del lenguaje natural (NLP). Tras los prometedores resultados obtenidos en dicho campo, varios autores comenzaron a explorar la posibilidad de aplicar esta arquitectura en otros dominios, como la visión por computador.

Desde su introducción, se ha demostrado que el rendimiento de los modelos \textit{Vision Transformer} (ViT) es comparable al de arquitecturas especializadas, como las redes convolucionales, en tareas clásicas de visión por computador tales como la clasificación de imágenes \cite{dosovitskiy2020image}, detección de objetos \cite{zhu2020deformable}, segmentación semántica \cite{zheng2021rethinking}, colorización de imágenes \cite{kumar2021colorization}, visión de bajo nivel \cite{chen2021pre} y comprensión de vídeo \cite{arnab2021vivit}, entre otras. Además, investigaciones recientes sugieren que los errores de predicción de los ViT son más coherentes con los de los humanos que los producidos por redes convolucionales \cite{tuli2021convolutional}.

El funcionamiento de los ViT es análogo al de un transformador convencional, siendo la principal diferencia la naturaleza de la señal de entrada. Mientras que en NLP se trabaja con secuencias de palabras, en visión se trabaja con imágenes. El desafío principal radica, por tanto, en adaptar el \textit{tokenizador} para procesar imágenes de forma compatible con la arquitectura del transformador.

Una de las primeras propuestas exitosas consiste en dividir la imagen en fragmentos disjuntos de tamaño fijo (por ejemplo, de $16 \times 16$ píxeles, como se plantea en \cite{dosovitskiy2020image}). Cada fragmento se aplanada y se proyecta a un espacio de dimensión fija mediante una capa lineal, generando así un vector que actúa como un \textit{token}. A este vector se le suma un \textit{embedding} posicional que codifica su localización dentro de la imagen, de forma análoga a cómo se representa la posición de palabras en secuencias de texto.

De este modo, se transforma la imagen original en una secuencia de tokens visuales que puede ser procesada por un transformador estándar, habilitando su uso en tareas propias de la visión por computador.

\begin{figure}[ht!]
	\centering
	\includegraphics[scale=0.5]{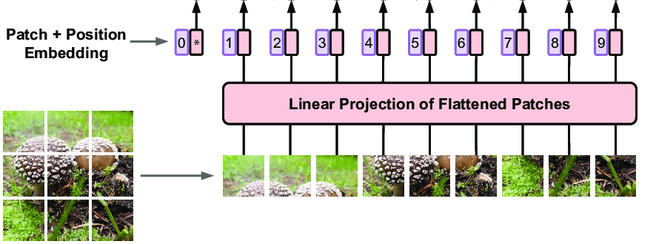} 
	\label{fig:embedding-imagen-transformador}
	\caption{Diagrama del proceso de partición, secuenciación y embedding de una imagen para su procesado en el transformador de visión (adaptado de \cite{picek2022automatic})}
\end{figure}

En \cite{liu2021survey} se puede encontrar un compendio exhaustivo de las referencias más relevantes hasta la fecha en el ámbito de los transformadores aplicados a visión por computador. Si bien el objetivo inicial de esta arquitectura es ofrecer un tratamiento generalista y unificado para datos heterogéneos, no todas las propuestas publicadas persiguen este mismo fin.

Así, es posible encontrar trabajos que modifican los mecanismos de atención para incorporar propiedades de localidad similares a las redes convolucionales, mediante estrategias como la atención local o jerárquica. Estos conceptos serán abordados con mayor detalle en el capítulo dedicado a los transformadores multimodales. Por otro lado, algunas aproximaciones utilizan como entrada al transformador representaciones derivadas de redes convolucionales preentrenadas, lo cual introduce un sesgo específico de dominio que se aleja del enfoque puramente generalista.

Aunque este módulo no pretende presentar la extensa variedad de variantes propuestas, muchas de las cuales se distancian del espíritu original de la arquitectura, se ofrecerá una visión general de las principales tendencias desarrolladas hasta el momento. Es importante destacar que los transformadores, tanto en general como en su aplicación a visión por computador, constituyen una arquitectura todavía joven y en evolución, por lo que cabe esperar importantes innovaciones en el futuro.

Más allá del interés que puedan suscitar aplicaciones concretas, el enfoque de este módulo se centrará en aquellas arquitecturas de carácter generalista que ofrecen buen rendimiento en un amplio espectro de tareas. En las secciones siguientes se presentarán las propuestas más representativas para las aplicaciones de visión por computador en clasificación, detección y segmentación. Para tareas más específicas, se remite al lector a la abundante bibliografía disponible sobre el tema.

\subsection{Clasificación}

Hasta el momento, se pueden identificar al menos seis enfoques distintos para abordar la aplicación de transformadores a tareas de clasificación en visión por computador. A continuación, se enumeran las principales estrategias descritas en la literatura:

\begin{enumerate}
	
	\item \textit{Adaptación directa de la arquitectura original} \cite{dosovitskiy2020image}. La imagen se divide en un conjunto de regiones no solapadas (\textit{patches}) que son tratadas como elementos secuenciales y procesadas mediante un transformador, en analogía con las secuencias de texto en NLP.
	
	\item \textit{Transformador de atributos procedentes de una red convolucional} \cite{wu2020visual}. En este enfoque, se extraen atributos de capas intermedias de una red convolucional preentrenada (CNN) y se utilizan como entrada de un transformador. Esta estrategia permite aprovechar los inductores estructurales de las CNN, y refinar las representaciones mediante la atención global de los transformadores.
	
	\item \textit{Destilación del conocimiento de una red convolucional preentrenada} \cite{touvron2021training}. Una CNN actúa como red maestra durante el entrenamiento de un modelo basado en transformadores, transfiriendo su conocimiento mediante técnicas de \textit{knowledge distillation} \cite{hinton2015distilling}. De forma notable, el modelo basado en transformadores puede llegar a superar el rendimiento de la red maestra de la que aprende.
	
	\item \textit{Transformadores con atención localizada y transformadores jerárquicos} \cite{liu2021survey}. Estos modelos modifican el mecanismo de atención para limitarlo a regiones locales de la imagen, en lugar de operar sobre toda la secuencia. A partir de una partición más fina de la imagen (por ejemplo, 4$\times$4 en lugar de 16$\times$16 como en \cite{dosovitskiy2020image}), las atenciones se restringen a grupos de regiones vecinas. Los resultados intermedios se fusionan progresivamente para ampliar el contexto, siguiendo una estructura piramidal o jerárquica. Trabajos como \cite{yuan2021tokens, wang2021pyramid} exploran variantes de esta idea, incorporando una reducción progresiva del espacio computacional a lo largo de las capas.
	
\end{enumerate}

\begin{figure}[ht!]
	\centering
	\includegraphics[width=0.8\textwidth]{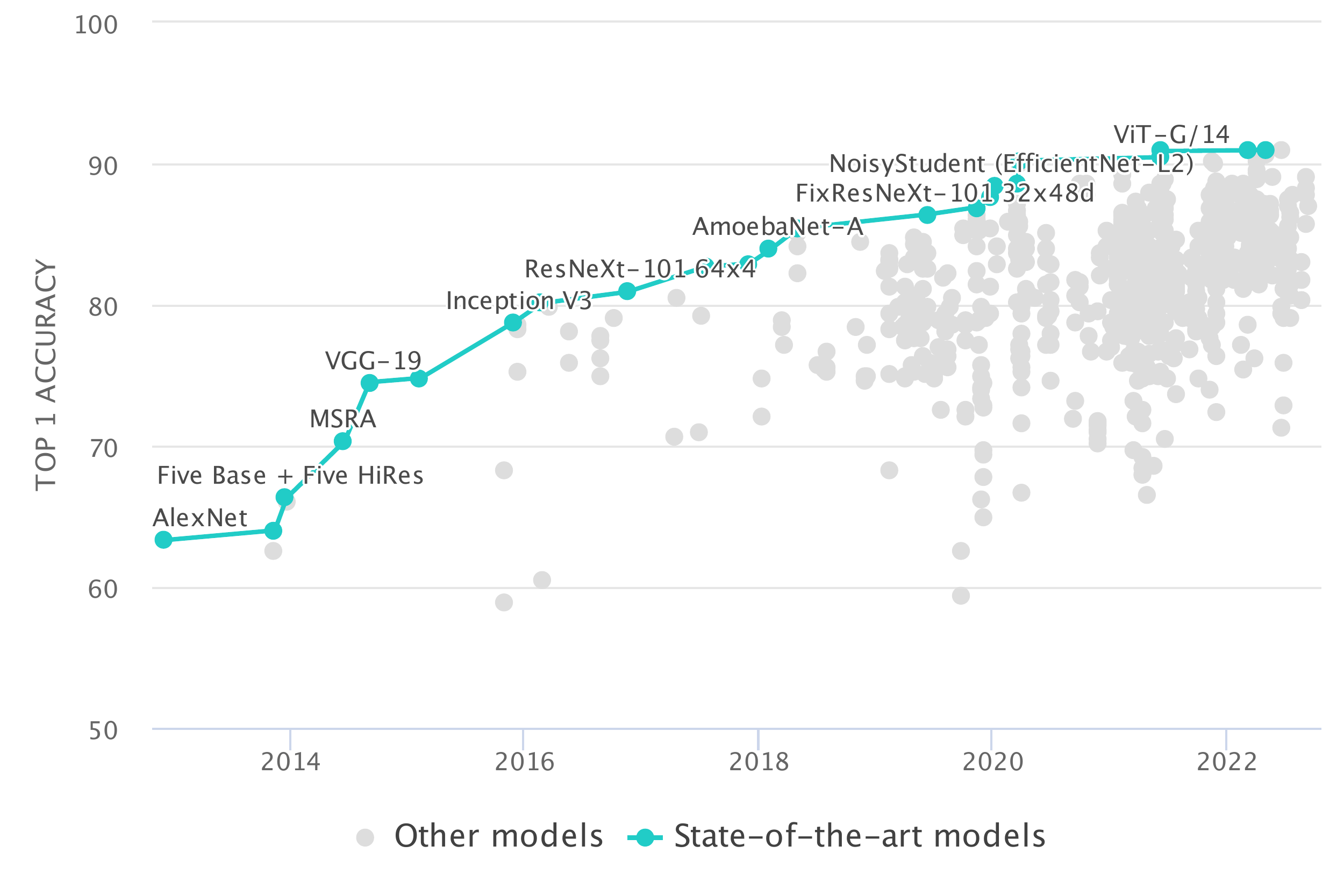} 
	\label{fig:imagenetclassificationleaderboard}
	\caption{Estado del arte de los modelos de clasificación de imagen \cite{imagenetclassificationleaderboard})}
\end{figure}

En la Figura~\ref{fig:imagenetclassificationleaderboard} se muestra la evolución temporal del estado del arte en clasificación de imágenes, utilizando como referencia el \textit{benchmark} ImageNet. Se observa que, hacia mediados de 2022, los modelos con mejor rendimiento están basados en transformadores, destacando especialmente \textbf{ViT-G/14}, con 1843 millones de parámetros y una precisión del $91.0\%$ \cite{ding2022davit}. Le sigue muy de cerca un modelo basado en redes convolucionales, \textbf{EfficientNet-L2}, con 480 millones de parámetros y un rendimiento del $90.2\%$ \cite{pham2021meta}. 

Para facilitar la comparación, se considera un modelo basado en transformadores con un número de parámetros similar al de EfficientNet-L2. En este caso, \textbf{DaViT-H}, con 362 millones de parámetros, también alcanza un rendimiento del $90.2\%$ \cite{ding2022davit}. A partir de estos resultados, puede concluirse que la arquitectura de los transformadores, a pesar de no incorporar inductores estructurales como las CNN, es capaz de obtener rendimientos comparables, e incluso superiores, con una complejidad computacional semejante.

Como contrapartida, cabe señalar que los modelos basados en transformadores requieren procesos de entrenamiento más prolongados y conjuntos de datos de gran tamaño para alcanzar dicho rendimiento. Una alternativa viable en contextos donde no se dispone de suficientes datos etiquetados es el uso de técnicas de aprendizaje semi-supervisado \cite{weng2021semi}.

\subsection{Detección}

Al referirnos al uso de transformadores para tareas de detección en imágenes, es necesario distinguir, desde un principio, entre aquellas aplicaciones que emplean un enfoque de transformador de extremo a extremo, y aquellas que lo utilizan únicamente como \textit{backbone} para la extracción de atributos, complementado con un bloque final de detección convencional (como por ejemplo \textbf{RetinaNet} \cite{lin2017focal} o \textbf{Mask R-CNN} \cite{he2017mask}).

Los \textit{backbones} mencionados suelen ser transformadores preentrenados en tareas de clasificación, reutilizados como extractores de atributos que alimentan un cabezal de detección tradicional. Para un compendio de los modelos que adoptan este tipo de enfoque, se puede consultar la bibliografía en \cite{liu2021survey}.

En cuanto a los modelos de detección basados en transformadores de extremo a extremo, al momento de redactar este capítulo existen principalmente dos paradigmas distintos para abordar el problema: por un lado, la familia de modelos derivados de la arquitectura \textbf{DETR} \cite{carion2020end}, y por otro, los modelos basados en la propuesta \textbf{Pix2Seq} \cite{chen2021pix2seq}. A continuación, se analizan las características definitorias de cada uno de ellos.

\subsubsection{\textbf{DETR}: \textbf{DE}tection with \textbf{TR}ansformer}

\begin{figure}[ht!]
	\centering
	\includegraphics[width=0.75\textwidth]{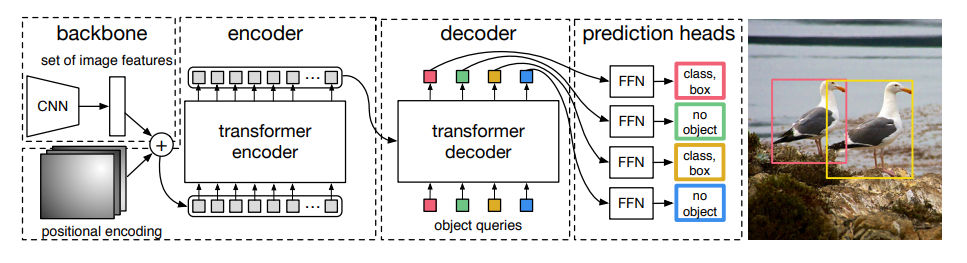} 
	\label{fig:detr}
	\caption{Arquitectura del transformador \textbf{DETR} especializado en detección \cite{carion2020end})}
\end{figure}

En la Figura~\ref{fig:detr} se muestra esquemáticamente la arquitectura de detección \textbf{DETR}. La imagen de entrada se divide en parches, se transforma en una secuencia y se introduce en un transformador de tipo \textit{secuencia a secuencia}. El \textit{tokenizador} está compuesto por una red convolucional que extrae atributos representativos de la imagen original, los cuales se utilizan como entrada del codificador del transformador.

La originalidad de esta propuesta radica en el diseño del decodificador, que recibe, por un lado, la salida del codificador, y por otro, una secuencia de entrada que representa propuestas iniciales de los objetos a detectar. A lo largo del decodificador, dicha secuencia se enriquece con la información contextual proveniente del codificador, generando una secuencia de salida que contiene las predicciones finales.

Si la red funciona correctamente, la secuencia de salida contendrá representaciones internas que codifican las detecciones de los objetos presentes en la imagen. Esta secuencia tiene la misma longitud que la secuencia de entrada, y cada uno de sus elementos se conecta a una red neuronal independiente encargada de predecir la clase y el cuadro delimitador correspondiente a cada posible detección. Por tanto, el tamaño de la secuencia del decodificador determina el número máximo de objetos que el modelo puede detectar. En el artículo original, este valor se fija en $100$. Para aquellas posiciones de la secuencia que no se corresponden con ningún objeto real, el modelo asigna una clase especial que indica la ausencia de detección.

En \textbf{DETR}, la secuencia de entrada del decodificador está compuesta por vectores aleatorios que actúan como \textit{priores} para la detección. A pesar de su naturaleza aleatoria, los resultados obtenidos son altamente competitivos. Trabajos posteriores han intentado mejorar las capacidades predictivas de la arquitectura afinando la generación de estos \textit{priores} en la secuencia de entrada, como es el caso de \textbf{SMCA}~\cite{gao2021fast}, \textbf{Conditional DETR}~\cite{meng2021conditional}, \textbf{Anchor DETR}~\cite{wang2022anchor}, \textbf{DAB-DETR}~\cite{liu2022dab}, \textbf{Efficient DETR}~\cite{yao2021efficient} y \textbf{Dynamic DETR}~\cite{dai2021dynamic}.

Un aspecto clave en el diseño de esta arquitectura es la definición de una función de pérdida que permita una optimización eficiente durante el aprendizaje supervisado, es decir, al comparar las predicciones del modelo con las etiquetas reales.

Para comprender la complejidad del problema, puede considerarse el ejemplo mostrado en la Tabla~\ref{table:deteccion-ejemplo-entrada}. En este se observa que las clases predichas y las reales no necesariamente ocupan las mismas posiciones dentro de la secuencia de salida. Además, el orden de las predicciones no garantiza una correspondencia directa con las etiquetas verdaderas, ya que es habitual que el modelo no detecte todos los objetos presentes o que lo haga de manera errónea.

Este problema se aborda evaluando la función de pérdida para todas las permutaciones posibles entre las predicciones y las etiquetas reales, seleccionando aquella asignación que minimiza el valor total de la pérdida. Esta formulación se inspira en trabajos previos como~\cite{carion2020end} y~\cite{stewart2016end}.

\begin{table}[ht]
	\centering
	\begin{tabular}{ | c | c | c | c | c | c | }
		\hline
		\multicolumn{3}{|c|}{\textbf{Real}} & \multicolumn{3}{c|}{\textbf{Predicho}} \\
		\hline
		\textbf{Clase} & \textbf{Desde} & \textbf{Hasta} & \textbf{Clase} & \textbf{Desde} & \textbf{Hasta} \\
		\hline
		vaso & (100, 200) & (200, 300) & sin objeto & (20, 40) & (10, 0) \\
		\hline
		botella & (300, 200) & (400, 300) & vaso & (98, 198) & (210, 310) \\
		\hline
		bol & (100, 200) & (200, 300) & sin objeto & (20, 40) & (10, 0) \\
		\hline
		portátil & (100, 200) & (200, 300) & sin objeto & (20, 40) & (10, 0) \\
		\hline
		ratón & (100, 200) & (200, 300) & sin objeto & (20, 40) & (10, 0) \\
		\hline
		silla & (500, 400) & (700, 500) & bolsa & (460, 560) & (510, 599) \\
		\hline
		bolsa & (450, 550) & (500, 600) & sin objeto & (20, 40) & (10, 0) \\
		\hline
		sin objeto & (0, 0) & (0, 0) & sin objeto & (20, 40) & (10, 0) \\
		\hline
		sin objeto & (0, 0) & (0, 0) & botella & (20, 40) & (10, 0) \\
		\hline
		sin objeto & (0, 0) & (0, 0) & sin objeto & (20, 40) & (10, 0) \\
		\hline
	\end{tabular}
	\caption{Ejemplo de valores reales y predichos en una red de detección}
	\label{table:deteccion-ejemplo-entrada}
\end{table}

Una vez realizado el emparejamiento óptimo, se emplea la función de pérdida húngara, mostrada en la Ecuación~\ref{eq:hungarian-loss}, para optimizar los parámetros de la red.

El emparejamiento óptimo se define como:

\begin{equation}
	\hat{\sigma} = \underset{\sigma \in \mathcal{G}_N}{\arg\min} \sum_{i=1}^{N} \mathcal{L}_{\text{match}}(y_i, \hat{y}_{\sigma(i)}),
	\label{eq:detr-loss}
\end{equation}

donde $\mathcal{G}_N$ denota el conjunto de todas las permutaciones posibles de $N$ elementos, y $\mathcal{L}_{\text{match}}$ es la pérdida de correspondencia definida como:

\begin{equation}
	\mathcal{L}_{\text{match}}(y_i, \hat{y}_{\sigma(i)}) = - \mathbb{1}_{\{c_i \ne \emptyset\}} \log \hat{p}_{\sigma(i)}(c_i) + \mathbb{1}_{\{c_i \ne \emptyset\}} \mathcal{L}_{\text{box}}(b_i, \hat{b}_{\sigma(i)}).
\end{equation}

Una vez determinada la asignación óptima $\hat{\sigma}$, la pérdida total utilizada para entrenar la red se define como:

\begin{equation}
	\mathcal{L}_{\text{H}} = \sum_{i=1}^{N} \left[ - \log \hat{p}_{\hat{\sigma}(i)}(c_i) + \mathbb{1}_{\{c_i \ne \emptyset\}} \mathcal{L}_{\text{box}}(b_i, \hat{b}_{\hat{\sigma}(i)}) \right],
	\label{eq:hungarian-loss}
\end{equation}

donde $\hat{p}_{\hat{\sigma}(i)}(c_i)$ representa la probabilidad predicha para la clase $c_i$ en la posición emparejada $\hat{\sigma}(i)$, y $\mathcal{L}_{\text{box}}$ corresponde a la pérdida de regresión del cuadro delimitador (por ejemplo, una combinación de L1 y GIoU).

\subsubsection{\textbf{Pix2Seq}}

\begin{figure}[ht!]
	\centering
	\includegraphics[width=0.6\textwidth]{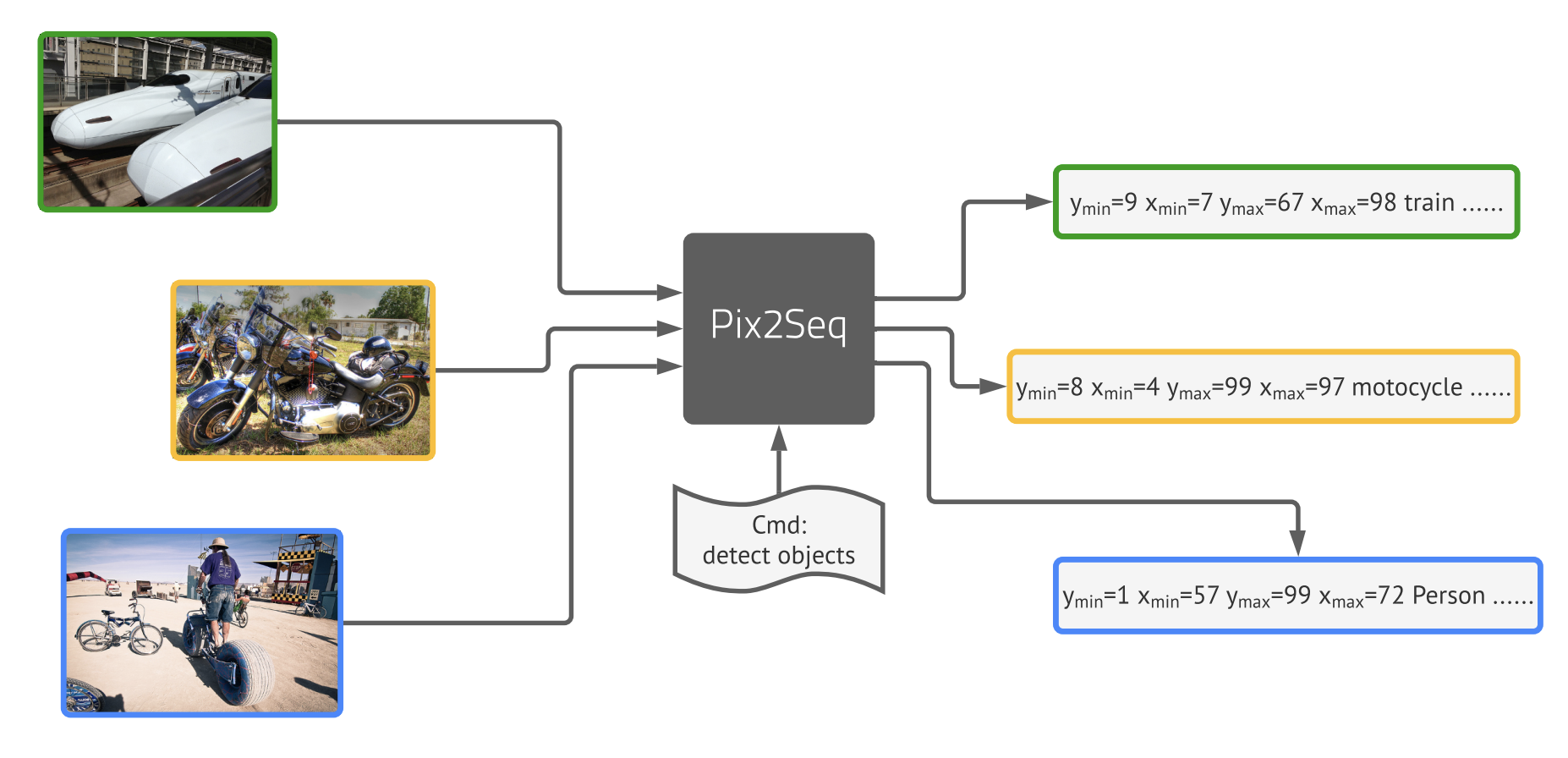} 
	\label{fig:detection-pix2seq}
	\caption{Tipología de E/S del transformador \textbf{Pix2Seq} especializado en detección \cite{carion2020end})}
\end{figure}

En la Figura~\ref{fig:detection-pix2seq} se muestra un esquema del funcionamiento de \textbf{Pix2Seq}, un nuevo paradigma que plantea un mapeo directo entre la imagen de entrada y una secuencia de texto como salida. En dicha secuencia se codifican tanto las clases como los cuadros delimitadores (bounding boxes) de todos los objetos detectados. Esta propuesta constituye una solución elegante, que ofrece un rendimiento comparable al de \textbf{DETR}, e incluso superior en ciertos escenarios, como en la detección de objetos pequeños.

El modelo se basa en una arquitectura de transformadores compuesta por un codificador y un decodificador que operan de manera auto-regresiva. La fase de codificación puede implementarse mediante un transformador o una red convolucional, cuya función es extraer los atributos relevantes de la imagen. Estos atributos se utilizan como entrada para el decodificador, el cual, junto con los tokens generados hasta el momento, produce la salida textual de forma secuencial. Cada elemento generado está condicionado tanto por las representaciones extraídas por el codificador como por los elementos previos de la secuencia.

La naturaleza textual de la salida simplifica los cálculos asociados a la función de pérdida en comparación con \textbf{DETR}. El formato de salida esperado consiste en una secuencia de elementos con la estructura: [x\_min], [y\_min], [x\_max], [y\_max], [clase], \dots, [\texttt{EOS}], donde \texttt{EOS} indica el final de la secuencia, y aparece tras la generación de todos los cuadros delimitadores presentes en la imagen.

Durante el entrenamiento, el orden de los elementos en la secuencia de predicción se establece de forma aleatoria. La optimización se realiza utilizando la función de pérdida habitual basada en \emph{log-likelihood}, empleada por los modelos auto-regresivos de procesamiento de lenguaje natural (NLP), como se muestra en la Ecuación~\ref{eq:loglikelihood}.

\begin{equation}
	\max \sum\limits_{j=1}^L \boldsymbol{\omega}_j \log P(\tilde{\boldsymbol{y}}_j \mid \boldsymbol{x}, \boldsymbol{y}_{1:j-1})
	\label{eq:loglikelihood}
\end{equation}

donde $\boldsymbol{x}$ representa la imagen de entrada, mientras que $\boldsymbol{y}$ y $\tilde{\boldsymbol{y}}$ corresponden a las secuencias de entrada y de salida objetivo asociadas a dicha imagen, respectivamente. El parámetro $L$ denota la longitud de la secuencia, y $\boldsymbol{\omega}_j$ es un peso predefinido asignado al elemento $j$-ésimo de la secuencia. En el caso particular de \textbf{Pix2Seq}, se utiliza un peso uniforme, es decir, $\boldsymbol{\omega}_j = 1\ \forall j$.

\subsection{Segmentación}

El término \textit{segmentación} se utiliza de forma general para englobar tres problemas diferentes: \textit{segmentación semántica}, \textit{segmentación de instancias} y \textit{segmentación panóptica}.

La \textit{segmentación semántica} consiste en asignar una clase a cada píxel de la imagen, sin distinguir entre distintas instancias de una misma clase. Se trata del tipo de segmentación más sencillo, ya que no requiere identificar objetos individuales.

Por otro lado, tanto la \textit{segmentación de instancias} como la \textit{segmentación panóptica} abordan un problema más complejo: no solo buscan clasificar los píxeles, sino también separar las distintas instancias de los objetos. La principal diferencia entre ambas radica en el tratamiento de la superposición: en la segmentación panóptica, cada píxel está asignado de forma única a una clase e instancia, mientras que en la segmentación de instancias se permite que un mismo píxel esté asociado a múltiples clases u objetos, lo que resulta útil en escenarios con solapamiento.

Las propuestas basadas en transformadores para abordar los problemas de segmentación han experimentado ---y continúan experimentando--- una profunda evolución. En general, pueden distinguirse dos enfoques principales: por un lado, la segmentación directa a partir de la información contenida en los píxeles de la imagen; y por otro, la construcción de representaciones internas centradas en objetos, a partir de las cuales se infieren posteriormente las máscaras de segmentación.

Las propuestas iniciales basadas en el método directo son una extensión del ViT para la resolución de tareas de segmentación (véase \textbf{SETR}~\cite{zheng2021rethinking}). Esta primera aproximación sirvió para demostrar la viabilidad de la arquitectura para la resolución de este tipo de problemas, aunque con unos costes computacionales elevados.

Posteriormente, se presentaron propuestas basadas en la representación interna de objetos con arquitecturas derivadas de la arquitectura de detección \textbf{DETR}, presentada en la sección anterior. Las primeras propuestas requerían de clase, cuadro delimitador y máscara. Las posteriores son capaces de derivar máscara y clase directamente, sin la necesidad de cuadro delimitador.

Presentaremos en esta sección la arquitectura \textbf{Mask2Former}~\cite{cheng2022masked}, como un ejemplo de una arquitectura madura que permite resolver con una única arquitectura las tres tareas de segmentación: semántica, de instancias y panóptica.

\begin{figure}[ht!]
	\centering
	\includegraphics[width=0.5\textwidth]{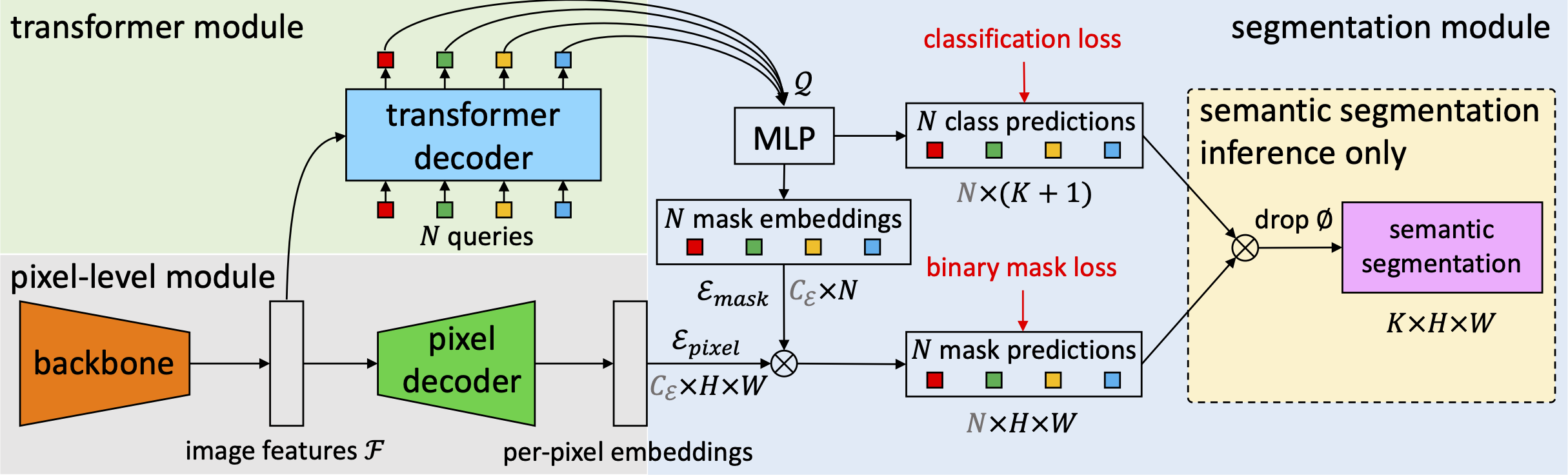} 
	\label{fig:segmentation-maskformer}
	\caption{Transformador MaskFormer. Arquitectura especializada en segmentación de imágenes. \cite{cheng2021per})}
\end{figure}

La Figura~\ref{fig:segmentation-maskformer} muestra un esquema representativo del transformador de segmentación \textbf{MaskFormer}, aplicable también a su versión mejorada, \textbf{Mask2Former}.

Podemos observar que la arquitectura consta de un \textit{backbone}, que puede ser tanto una red convolucional como un ViT, a partir del cual se extraen los atributos representativos de la imagen. Estos atributos se alimentan a dos módulos principales: un decodificador de píxeles y un decodificador basado en transformadores.

El decodificador de píxeles está implementado mediante una \textit{Feature Pyramid Network} (FPN)~\cite{lin2017feature}, aunque podría sustituirse por cualquier otra arquitectura capaz de transformar representaciones de atributos en mapas espaciales, como por ejemplo una \textit{U-Net}~\cite{ronneberger2015u}.

Por otro lado, el decodificador de transformadores, siguiendo el enfoque de \textbf{DETR}, genera una secuencia de salida a partir de una secuencia de entrada, utilizando como base la representación interna de los objetos extraída por el \textit{backbone}. A partir de esta representación, un MLP produce la predicción de clase para cada objeto, así como una representación latente de la máscara correspondiente.

Esta representación de máscara se combina con la salida del decodificador de píxeles para generar las máscaras finales de segmentación para cada objeto. Al igual que en \textbf{DETR}, se incluye una clase especial que permite identificar aquellas predicciones donde no se ha detectado ningún objeto, y cuyas máscaras asociadas deben ser descartadas.

El número máximo de máscaras que la red puede predecir es un valor de diseño fijado por el tamaño de la secuencia de objetos procesada por el decodificador de transformadores.

\section{Transformadores para Audio}

Los transformadores se están consolidando como el estándar \textit{de facto} para el procesamiento de señales de audio, un dominio que hasta hace poco estaba dominado por redes convolucionales y, anteriormente, por redes recurrentes. Las posibilidades de aplicación en este campo son amplias y diversas. La arquitectura empleada es, en esencia, la misma que en otras aplicaciones de transformadores para distintos tipos de señales; el elemento distintivo radica en el \textit{tokenizador} utilizado, encargado de transformar la señal de entrada en una representación interna adecuada para ser procesada por la red.

\subsection{Señales de audio: tratamiento y tokenización}

Las entradas de texto e imagen resultan, por lo general, más intuitivas de comprender. Sin embargo, el tratamiento del sonido presenta una mayor complejidad en cuanto a su representación numérica. En esta sección se describe la naturaleza de las señales de audio con el objetivo de entender mejor cómo se preparan para su uso en tareas de predicción automática.

Una señal puede entenderse como la evolución temporal de una magnitud física medible. En el caso del audio, se refiere a las variaciones de presión del aire en un punto determinado. Para poder tratar estas señales de forma digital, el primer paso consiste en almacenarlas. Esto se logra mediante el \textit{muestreo}, que implica registrar el valor de la señal a intervalos regulares según una frecuencia de muestreo preestablecida. Un valor habitual para señales de sonido es de $\SI{44.1}{kHz}$, es decir, $44100$ muestras por segundo. Una vez muestreada, la señal de audio queda representada digitalmente y puede ser procesada mediante técnicas computacionales.

Desde el análisis matemático, el teorema de Fourier establece que cualquier señal periódica definida en un intervalo temporal finito puede descomponerse como suma de señales senoidales, cada una caracterizada por una frecuencia y una amplitud. Estas componentes pueden representarse gráficamente en un \textit{espectro de frecuencias}, lo que permite describir una misma señal tanto en el dominio temporal como en el dominio frecuencial. La transformación entre ambos dominios se denomina \textit{Transformada de Fourier} \cite{bracewell1986fourier}, y su implementación eficiente en el entorno digital se logra mediante el algoritmo FFT (\textit{Fast Fourier Transform}) \cite{nussbaumer1981fast}.

No obstante, el habla y la música no son señales periódicas. En estos casos, se emplea una técnica conocida como \textit{ventaneo}, en la que la señal se divide en segmentos temporales de longitud fija en los que se asume periodicidad local. Para cada ventana se calcula su espectro de frecuencias, y al superponer sucesivamente estas ventanas se obtiene una representación frecuencial dependiente del tiempo: el \textit{espectrograma}. En este, el eje horizontal representa el tiempo, el eje vertical la frecuencia, y la intensidad de cada punto se codifica mediante una escala de colores que indica la amplitud (habitualmente en decibelios, es decir, en escala logarítmica). Esta escala se emplea por estar más alineada con la forma en que los humanos percibimos los cambios de volumen.

De manera similar, la percepción humana de las frecuencias tampoco es lineal. El oído humano es más sensible a los cambios en frecuencias bajas que en altas. Por ejemplo, una persona puede distinguir fácilmente una señal de $\SI{500}{Hz}$ de una de $\SI{1000}{Hz}$, pero tendrá mayores dificultades para percibir la diferencia entre una de $\SI{10000}{Hz}$ y una de $\SI{10500}{Hz}$, a pesar de que la diferencia absoluta es la misma. Para representar esta no linealidad perceptiva, en 1937 se propuso la \textit{escala MEL} \cite{stevens1937scale}, que introduce una nueva unidad, el \textit{pitch} MEL, para aproximar linealmente la percepción de las diferencias de frecuencia. El gráfico que resulta de representar la amplitud (en decibelios) sobre un plano \textit{pitch MEL} frente a tiempo se conoce como \textit{espectrograma MEL}, y es una representación más fiel a la percepción auditiva humana. Por esta razón, el espectrograma MEL se ha convertido en la representación más común en aplicaciones de aprendizaje automático con audio, ya que resalta las características más relevantes para el reconocimiento y facilita su posterior procesamiento.
 
Una vez obtenida esta representación en forma de matriz bidimensional, su tamaño suele exceder la capacidad de entrada directa de un transformador. Por ello, un \textit{tokenizador} habitual consiste en aplicar capas convolucionales para reducir dimensionalmente el espectrograma MEL, adaptándolo así al espacio interno de representación de la arquitectura transformadora. Cabe destacar que también existen otros métodos de tokenización alternativos, dependiendo del tipo de tarea y arquitectura empleada.

\subsection{Ejemplos de transformadores aplicados a audio}

A continuación, se presentan algunos ejemplos representativos de modelos basados en arquitecturas de transformadores aplicados al procesamiento de señales de audio. Estos modelos ilustran la diversidad de tareas que pueden abordarse en este dominio:

\begin{itemize}
	\item \textbf{Whisper} \cite{radfordrobust} es un modelo de tipo \textit{secuencia a secuencia} multilingüe diseñado para la transcripción automática de audio a texto. A partir de una señal de entrada que contiene una conversación (por ejemplo, un archivo de voz), el modelo devuelve la secuencia de texto correspondiente. Whisper ha sido entrenado con una gran cantidad de datos procedentes de múltiples idiomas, lo que le confiere una notable robustez en condiciones de ruido y variabilidad del habla.
	
	\item \textbf{Wav2Vec 2.0} \cite{baevski2020wav2vec} es un modelo autoregresivo para el aprendizaje de representaciones de audio sin supervisión. Su arquitectura combina una red convolucional que actúa como codificador de características (feature encoder) con un transformador que modela dependencias de largo alcance. Al igual que BERT en el procesamiento de texto, Wav2Vec 2.0 utiliza una tarea de enmascaramiento para aprender representaciones contextuales que pueden ser luego ajustadas (\textit{fine-tuned}) con un número reducido de ejemplos etiquetados para tareas como el reconocimiento automático del habla (ASR).
	
	\item \textbf{SepFormer} \cite{subakan2021attention} es un modelo basado en transformadores desarrollado para la tarea de separación de fuentes, especialmente la separación de voces en grabaciones con múltiples interlocutores. Su arquitectura explota mecanismos de atención para modelar relaciones temporales a largo plazo, superando en rendimiento a modelos previos tanto en entornos limpios como ruidosos.
\end{itemize}

\section{Transformadores Multi-Modales}

Hasta ahora hemos explorado el potencial de los transformadores para el aprendizaje automático utilizando datos de una única naturaleza. Dada su capacidad para abordar problemas de áreas previamente desconectadas, es razonable hipotetizar que esta arquitectura podría ser igualmente eficaz en la gestión de problemas que requieran la integración simultánea de entradas de distintas naturalezas. Un ejemplo claro de un agente con tales capacidades son los animales y las personas, que integran información proveniente de hasta cinco sentidos diferentes.

Las arquitecturas \textit{multi-modales} son aquellas capaces de gestionar entradas de diferentes naturalezas, como imágenes y texto. 

Desde el punto de vista de la geometría topológica, el espacio generado por la arquitectura de los transformadores es equivalente a un grafo completamente conectado \cite{bronstein2021geometric}. Este comportamiento contrasta con otras arquitecturas, donde las interconexiones del grafo están limitadas a un espacio más restringido. En consecuencia, los transformadores disponen de un espacio de modelado más general y flexible \cite{xu2022multimodal}, lo que representa una ventaja cuando se requiere combinar entradas de distinta naturaleza. Sin embargo, los priors estructurales que pueden ser útiles para facilitar la optimización en un tipo de datos específico pueden convertirse en un inconveniente cuando se usan junto con otras entradas de diferente naturaleza.

A continuación, se describen los puntos diferenciales entre una arquitectura \textit{multi-modal} y una tradicional.

\subsection{Entradas Multi-Modales}

La primera función que realiza un transformador es la conversión de los datos de entrada en un tensor de dimensión interna definida, que se mantiene constante a medida que avanza a través de las distintas capas. La \textit{tokenización} transforma los elementos de entrada en un vector intermedio al que posteriormente se le aplica una transformación matemática, que normalmente reduce sus dimensiones con el objetivo de alcanzar las dimensiones internas de proceso. 

En el caso de las entradas textuales, es habitual utilizar un \textit{tokenizador} que convierte cada elemento de la cadena de entrada en un vector ortogonal de dimensión igual a la del diccionario. Esta operación es seguida de una transformación lineal con el objetivo de alcanzar la dimensión interna de trabajo. 

Para las imágenes, una estrategia común consiste en \textit{tokenizar} la imagen mediante su partición en partes no superpuestas de tamaño predefinido, seguida de una transformación lineal para alcanzar la dimensión de trabajo \cite{dosovitskiy2020image}. Otra estrategia de \textit{tokenización} implica el uso de redes especializadas para derivar atributos que se alimentan al transformador \cite{lu2019vilbert}. 

En el caso del video, existen diversas estrategias. En \cite{sun2019videobert}, la tokenización consiste en seleccionar clips de duración predeterminada a unos fps preestablecidos, a los cuales se les aplica una red convolucional tridimensional para generar atributos que luego se adaptan hasta conseguir las dimensiones internas de trabajo. Otras estrategias publicadas incluyen la tokenización mediante la selección de puntos tridimensionales de interés en los clips, seguida de una transformación lineal \cite{akbari2021vatt}, o la tokenización mediante la partición de las imágenes integrantes del clip, seguida de una proyección lineal \cite{nagrani2021attention}.

En el caso del audio, se pueden emplear diversas estrategias que dependen del objetivo específico que se pretenda alcanzar. Una de ellas consiste en utilizar el espectrograma MEL (u otros derivados), seguido de una proyección lineal o del cálculo de atributos mediante una red convolucional \cite{lin2021exploring}, \cite{akbari2021vatt}, \cite{nagrani2021attention}. 

En la bibliografía de \cite{xu2022multimodal} se pueden encontrar ejemplos de módulos preparatorios para otros tipos de datos más específicos, como esquemas de bases de datos SQL, nubes de puntos tridimensionales, datos tabulares, poses humanas, registros electrónicos de salud, conjuntos de órdenes posibles a los que responde un robot, entre otros muchos.

\subsection{Variantes de auto-atención en un contexto multi-modal}

Una vez que se dispone de la representación interna de las distintas señales de entrada, surge el dilema de cómo encontrar el modo más efectivo de combinarlas en la capa de auto-atención para obtener los mejores resultados. A pesar de que las señales puedan tener las mismas dimensiones, debido a su diferente naturaleza, puede ser necesario adaptar las transformaciones a cada uno de los distintos modos. 

Las estrategias más habituales para calcular la auto-atención en esta primera etapa son: la \textit{suma temprana}, la \textit{concatenación temprana}, la \textit{jerárquica} y la \textit{cruzada}. 

La \textit{suma temprana} (ver Fig. \ref{fig:att-sum}) consiste en sumar los vectores procedentes de las distintas fuentes como paso previo a un único sistema de auto-atención común.

\begin{figure}[ht!]
	\centering
	\includegraphics[width=0.3\textwidth]{figs/early\_summation\_attention.pdf} 
	\label{fig:att-sum}
	\caption{Atención por suma temprana. La información procedente de dos canales distintos se suma antes de ser procesada por el canal de auto-atención}
\end{figure}

La \textit{concatenación temprana} (fig. \ref{fig:att-concat}) genera un vector como concatenación de los distintos modos que se alimenta al módulo de auto-atención. 

\begin{figure}[ht!]
	\centering
	\includegraphics[width=0.6\textwidth]{figs/early\_concatenation\_attention.pdf} 
	\label{fig:att-concat}
	\caption{Atención por concatenación. La información procedente de dos canales distintos se procesa en la misma capa de auto-atención}
\end{figure}

La \textit{auto-atención jerárquica} (fig. \ref{fig:att-jer} y \ref{fig:att-jer2}) se realiza en dos pasos, en uno se aplican transformaciones distintas a cada señal y en el otro una única que las combina. 

\begin{figure}[ht!]
	\centering
	\includegraphics[width=0.6\textwidth]{figs/hierarchical\_attention.pdf} 
	\label{fig:att-jer}
	\caption{Atención jerárquica multi-canal a mono-canal. La información procedente de canales distintos se procesa en una primera etapa en canales de atención independientes para posteriomente procesarse en conjunto en la etapa final.}
\end{figure}

\begin{figure}[ht!]
	\centering
	\includegraphics[width=0.6\textwidth]{figs/hierarchical\_attention\_2.pdf} 
	\label{fig:att-jer2}
	\caption{Atención jerárquica mono-canal a multi-canal. La información procedente de canales distintos se procesa en una primera etapa en conjunto para posteriomente procesarse en la etapa final de manera independiente}
\end{figure}

La \textit{auto-atención cruzada} (fig. \ref{fig:cross-concat}) utiliza canales de auto-atención independientes donde la matriz de Query de los canales actúa sobre el otro canal y no sobre el propio. 

\begin{figure}[ht!]
	\centering
	\includegraphics[width=0.5\textwidth]{figs/cross\_attention.pdf} 
	\label{fig:cross-concat}
	\caption{Atención cruzada. La información procedente de dos canales distintos se procesa en la misma capa de auto-atención}
\end{figure}

La \textit{auto-atención cruzada seguida de concatenación} (fig. \ref{fig:cross-concat2}) actúa en dos fases, una primera en que se aplica una auto-atención cruzada y posteriormente se aplica una auto-atención convencional sobre la concatenación de los vectores procedentes de la operación original.

\begin{figure}[ht!]
	\centering
	\includegraphics[width=0.6\textwidth]{figs/cross\_attention2.pdf} 
	\label{fig:cross-concat2}
	\caption{Atención cruzada seguida de concatenación. La información procedente de dos canales distintos se procesa en la misma capa de auto-atención y posteriormente es combinada en un canal único}
\end{figure}

\begin{table}[ht!]
	\centering
	\begin{tabular}{ l | c | c}
			\hline %\rowcolor[gray]{0.8}
			%% header
			\textbf{Auto-atención} & \textbf{Formulación} & Referencias\\
			\hline
			Suma temprana & $\boldsymbol{Z} \leftarrow Tf(\alpha \boldsymbol{Z}_{(A)} \bigoplus \beta \boldsymbol{Z}_{(B)})$ & \cite{gavrilyuk2020actor}\\
			&&\cite{xu2021deepchange}\\
			\hline
			Concatenación temprana & $\boldsymbol{Z} \leftarrow Tf(\mathcal{C} (\boldsymbol{Z}_{(A)}, \boldsymbol{Z}_{(B)})$ & \cite{sun2019videobert}\\
			&&\cite{guo2020graphcodebert}\\
			&&\cite{shi2022learning}\\
			&&\cite{zheng2021fused}\\
			\hline
			Jerárquica multi $\rightarrow$ mono-canal & $\boldsymbol{Z} \leftarrow Tf_3(\mathcal{C}( Tf_1(\boldsymbol{Z}_{(A)}), Tf_2(\boldsymbol{Z}_{(B)}))$ & \cite{li2021ai}\\	
			\hline	
			Jerárquica mono $\rightarrow$ multi-canal & $\Bigg\{ \begin{array}{c} \boldsymbol{Z}_{(A)} \leftarrow Tf_2(Tf_1(\mathcal{C} (\boldsymbol{Z}_{(A)}, \boldsymbol{Z}_{(B)})))\\\boldsymbol{Z}_{(B)} \leftarrow Tf_3(Tf_1(\mathcal{C} (\boldsymbol{Z}_{(A)}, \boldsymbol{Z}_{(B)})))\end{array}$& \cite{lin2020interbert}\\	
			\hline	
			Cruzada & $\Bigg\{ \begin{array}{c} \boldsymbol{Z}_{(A)} \leftarrow MHSA(\boldsymbol{Q_B}, \boldsymbol{K_A}, \boldsymbol{V_A}) \\\boldsymbol{Z}_{(B)} \leftarrow MHSA(\boldsymbol{Q_A}, \boldsymbol{K_B}, \boldsymbol{V_B}) \end{array}$ & \cite{lu2019vilbert}\\
			&&\cite{yun2021pano}\\
			\hline
			Cruzada más concatenación & $\Bigg\{ \begin{array}{c} \boldsymbol{Z}_{(A)} \leftarrow MHSA(\boldsymbol{Q_B}, \boldsymbol{K_A}, \boldsymbol{V_A}) \\\boldsymbol{Z}_{(B)} \leftarrow MHSA(\boldsymbol{Q_A}, \boldsymbol{K_B}, \boldsymbol{V_B}) \\ \boldsymbol{Z} \leftarrow Tf(\mathcal{C}(\boldsymbol{Z}_{(A)}, \boldsymbol{Z}_{(B)})) \end{array}$&\cite{hasan2021humor}\\	
			&&\cite{zhan2021product1m}\\
			&&\cite{tsai2019multimodal}\\	
			\hline
		\end{tabular}
	\caption{Variantes de auto-atención de interacción/fusión de canales multi-modales. $\alpha$ y $\beta$ son pesos. Att: Atención; $\mathcal{C}$ : Concatenación; Tfs: Capas de transformador. $N_{(A)}$ y $N_{(B)}$ longitudes de secuencia de dos canales A y B de distinto modo. $MHSA(\boldsymbol{Q}, \boldsymbol{K}, \boldsymbol{V})$ Función de auto-atención multicabezal}
	\label{table:auto-atencion-variantes}
\end{table}

\subsection{Arquitecturas multimodales}

El objetivo de estas estrategias es obtener una representación interna que integre de la manera más adecuada posible la información relevante proveniente de cada uno de los modos de entrada. El sistema de auto-atención seleccionado determina las características de la arquitectura. En este sentido, podemos hablar de arquitecturas de \textit{canal único}, \textit{multicanal} o \textit{híbridas}.

Las arquitecturas de \textit{canal único} son aquellas derivadas del uso de sistemas de auto-atención basados en \textit{suma} y \textit{concatenación temprana}, y disponen de un único canal para el tratamiento de los datos. 

Las arquitecturas \textit{multicanal} se derivan del uso de sistemas de auto-atención \textit{cruzada}, lo que da lugar a múltiples canales de tratamiento diferenciado de los datos. 

Finalmente, las arquitecturas \textit{híbridas}, que surgen de la combinación de sistemas de auto-atención \textit{jerárquica} y \textit{cruzada con concatenación}, integran elementos de las dos arquitecturas anteriores.

\section{Aplicaciones de los transformadores multimodales}

En esta sección se presentan algunos de los trabajos más destacados publicados hasta el momento que emplean transformadores multimodales.

\begin{itemize}
	\item \textbf{MMBT} \cite{kiela2019supervised} es una arquitectura multimodal que integra imagen y texto para obtener una representación interna que puede utilizarse en tareas de clasificación.
	
	\item \textbf{VideoBERT} \cite{sun2019videobert} es una arquitectura multimodal similar a BERT, pero utilizando video como fuente de datos.
	
	\item \textbf{Med-BERT} \cite{rasmy2021med} es una arquitectura multimodal que emplea registros de salud estructurados para representar la salud de los pacientes y, a partir de esta representación, predecir enfermedades.
	
	\item \textbf{Gato} \cite{reed2022generalist} es una arquitectura multimodal generalista que, utilizando datos procedentes de diversas fuentes (imagen, texto y acciones propias de un robot), es capaz de resolver hasta 604 tareas diferentes. Entre estas tareas se incluyen juegos de Atari, chatbots, control de brazos robóticos y etiquetado de imágenes, entre muchas otras. Todo ello utilizando una única arquitectura basada en transformadores. Es la primera aplicación generalista que integra imagen, procesamiento del lenguaje natural, aprendizaje por refuerzo y control de robots en un solo agente.
\end{itemize}

%%%%%%%%
% RESUM %
%%%%%%%%
\newpage
\section{Resumen}

En este módulo, hemos presentado la arquitectura de redes neuronales de tipo \textit{Transformadores}, la cual tiene la capacidad de reemplazar a arquitecturas especializadas como redes convolucionales, recurrentes y de grafos. Esta arquitectura no solo ha sido diseñada originalmente para el procesamiento del lenguaje natural, sino que también ha demostrado un rendimiento excepcional en el tratamiento de otros tipos de señales específicas, así como en la resolución de problemas que combinan datos de distintas fuentes. Esto la convierte en un agente generalista para la resolución de problemas en diversas áreas.

La arquitectura de los Transformadores ha mostrado resultados sobresalientes en una variedad de dominios, tales como: visión por computador, procesamiento del lenguaje natural, tratamiento de señales de audio, predicción de series temporales y aprendizaje por refuerzo, entre otros.

En el módulo, se ha presentado una descripción rigurosa del algoritmo que subyace a esta arquitectura, mostrando aplicaciones en distintas áreas. Desde su aplicación original en el procesamiento del lenguaje natural, pasando por su uso en imágenes y sonidos, hasta su adopción en aplicaciones \textit{multi-modales}.

Conocer los detalles de implementación de la arquitectura de los Transformadores es crucial, ya que representa una solución general para una amplia gama de problemas en ciencia de datos.

\newpage
\section*{Notación}

La notación matemática de este módulo es la misma que la utilizada en \cite{phuong2022formal}. A efectos prácticos la reproducimos a continuación.

\begin{tabular}{ l  l  l }
	\hline %\rowcolor[gray]{0.8}
	%% header
	\textbf{Símbolo} & \textbf{Tipo} & \textbf{Explicación} \\
	\hline
	$[N]$ & $:= \lbrace 1,..., N\rbrace$ & conjunto de enteros $1, 2, ..., N - 1, N$ \\
	$i, j$ & $\in \mathbb{N}$ & índices enteros genéricos \\
	$V$ & $\cong [N_V]$ & vocabulario \\
	$N_V$ & $ \in \mathbb{N}$ & tamaño del vocabulario \\
	$V^*$ & $ = \bigcup_{\ell=0}^{\infty} V^{\ell}$ & conjunto de secuencias de tokens; p.e. palabras y documentos \\
	$\ell_{max}$ & $\in \mathbb{N}$ & longitud máxima de la secuencia de tokens\\
	$\ell$ & $\in [ \ell_{max} ] $ & longitud de la secuencia de tokens\\
	$t$ & $ \in [ \ell ] $ & índice del token dentro de una secuencia\\
	$d_{...}$ & $\in \mathbb{N}$ & dimensión de varios vectores\\
	$\mathbf{x}$ & $\equiv x[1 : \ell ]$ & $\equiv x[1]x[2]...x[\ell] \in V^\ell $ secuencia de tokens primaria\\
	$\mathbf{z}$ & $\equiv z[1 : \ell ]$ & $\equiv z[1]z[2]...z[\ell] \in V^\ell $ secuencia de tokens de contexto\\
	$M[i,j]$ & $ \in \mathbb{R}$ & entrada $M_{i,j}$ de la matriz $M \in \mathbb{R}^{d \times d'}$\\
	$M[i,:] \equiv M[i]$ & $ \in \mathbb{R}^{d'}$ & fila $i$ de la matriz $M \in \mathbb{R}^{d \times d'}$\\
	$M[:,j]$ & $ \in \mathbb{R}^{d}$ & columna $j$ de la matriz $M \in \mathbb{R}^{d \times d'}$\\
	$\mathbf{e}$ & $\mathbb{R}^{d_e}$ & representación vectorial / representación vectorial (embedding) de un token\\
	$\mathbf{X}$ & $\mathbb{R}^{d_e \times \ell_\mathcal{X}}$ & codificación de la secuencia de tokens primaria \\
	$\mathbf{Z}$ & $\mathbb{R}^{d_e \times \ell_\mathcal{Z}}$ & codificación de la secuencia de tokens de contexto \\
	$\mathbf{Mask}$& $\mathbb{R}^{\ell_\mathcal{Z} \times \ell_\mathcal{X}}$ & matriz de enmascarado, determina el contexto de atención para cada token\\
	$L, L_{enc}, L_{dec}$ & $\mathbb{N}$ & número de capas de red (codificación y decodificación)\\
	$l$&$\in [L]$ &índice de la capa de la red\\
	$H$ & $\mathbb{N}$ & número de cabezales de atención\\
	$h$ & $\in [H]$ & índice del cabezal de atención\\
	$N_{data}$ & $\in \mathbb{N}$& (i.i.d.) tamañp de la muestra\\
	$n$ & $\in [N_{data}]$ & índice de la secuencia de muestra\\
	$\eta$&$\in (0, \infty)$ & ratio de aprendizaje\\
	$\tau$&$\in (0, \infty)$ & temperatura, controla la diversidad en tiempo de inferencia\\
\end{tabular}

\begin{tabular}{ l  l  l }
	\hline %\rowcolor[gray]{0.8}
	%% header
	\textbf{Símbolo} & \textbf{Tipo} & \textbf{Explicación} \\
	\hline
	$\mathbf{W_e}$&$\in \mathbb{R}^{d_e \times N_V}$& matriz de embeddings de los tokens\\
	$\mathbf{W_p}$&$\in \mathbb{R}^{d_e \times \ell_{max}}$& matriz de embeddings de posición\\
	$\mathbf{W_u}$&$\in \mathbb{R}^{N_V \times d_e}$& matriz de conversión de embedding a token\\
	$\mathbf{W_q}$&$\in \mathbb{R}^{d_{attn} \times d_{\mathcal{X}}}$& parámetros de la matriz de consulta\\
	$\mathbf{b_q}$&$\in \mathbb{R}^{d_{attn}}$& sesgo de consulta\\
	$\mathbf{W_k}$&$\in \mathbb{R}^{d_{attn} \times d_{\mathcal{Z}}}$& parámetros de la matriz de clave\\
	$\mathbf{b_k}$&$\in \mathbb{R}^{d_{attn}}$& sesgo de clave\\
	$\mathbf{W_v}$&$\in \mathbb{R}^{d_{out} \times d_{\mathcal{Z}}}$& parámetros de la matriz de valor\\
	$\mathbf{b_v}$&$\in \mathbb{R}^{d_{attn}}$& sesgo de valor\\
	$\mathbf{W_{qkv}}$ & & colección de parámetros de una capa de atención \\
	$\mathbf{W_o}$&$\in \mathbb{R}^{d_{out} \times Hd_{mid}}$& parámetros de la matriz de salida\\
	$\mathbf{b_o}$&$\in \mathbb{R}^{d_{out}}$& sesgo de salida\\
	$\mathbf{W}$& & colección de parámetros de una capa de multi-atención\\
	$\mathbf{W_{mlp}}$&$\in \mathbb{R}^{d_1 \times d_2}$& parámetros de una MLP del transformador\\
	$\mathbf{b_{mlp}}$&$\in \mathbb{R}^{d_1}$& sesgo correspondiente a una MLP del transformador\\
	$\mathbf{\gamma}$&$\in \mathbb{R}^{d_e}$&parámetro de escalado de la capa de normalización\\
	$\mathbf{\beta}$&$\in \mathbb{R}^{d_e}$&parámetro de offset de la capa de normalización\\
	$\mathbf{\theta}, \mathbf{\hat{\theta}}$&$\in \mathbb{R}^{d}$&colección de todos los parámetros del transformador\\
\end{tabular}

\newpage

\nocite{*}
\bibliographystyle{unsrtnat}
\bibliography{transformadores.bbl}  %%% Uncomment this line and comment out the ``thebibliography'' section below to use the external .bib file (using bibtex) .

\end{document}